\documentclass[letterpaper]{article} % DO NOT CHANGE THIS
\usepackage{aaai25}  % DO NOT CHANGE THIS
\usepackage{times}  % DO NOT CHANGE THIS
\usepackage{helvet}  % DO NOT CHANGE THIS
\usepackage{courier}  % DO NOT CHANGE THIS
\usepackage[hyphens]{url}  % DO NOT CHANGE THIS
\usepackage{graphicx} % DO NOT CHANGE THIS
\usepackage{natbib}  % DO NOT CHANGE THIS AND DO NOT ADD ANY OPTIONS TO IT
\usepackage{caption} % DO NOT CHANGE THIS AND DO NOT ADD ANY OPTIONS TO IT
\usepackage{algorithm}
\usepackage{algorithmic}
\usepackage{amsmath,amsfonts}
\usepackage{subfigure}
\usepackage{booktabs}
\usepackage{pifont}
\usepackage{multirow}
\usepackage{newfloat}
\usepackage{listings}
\DeclareCaptionStyle{ruled}{labelfont=normalfont,labelsep=colon,strut=off} % DO NOT CHANGE THIS
\floatstyle{ruled}
\newfloat{listing}{tb}{lst}{}
\floatname{listing}{Listing}
\title{FNIN: A Fourier Neural Operator-based Numerical Integration Network for Surface-form-gradients}
\author {
    Jiaqi Leng\textsuperscript{\rm 1,2},
    Yakun Ju\textsuperscript{\rm 3}\thanks{Corresponding author.},
    Yuanxu Duan\textsuperscript{\rm 4},
    Jiangnan Zhang\textsuperscript{\rm 5},
    Qingxuan Lv\textsuperscript{\rm 5},\\
    Zuxuan Wu\textsuperscript{\rm 2},
    Hao Fan\textsuperscript{\rm 5}\footnotemark[1]
}
\affiliations {
    \textsuperscript{\rm 1}Haide College, Ocean University of China \\
    \textsuperscript{\rm 2}Shanghai Key Lab of Intell. Info. Processing, School of CS, Fudan University\\
    \textsuperscript{\rm 3}School of Computing and Mathematic Sciences, University of Leicester\\
     \textsuperscript{\rm 4}Department of Computer Engineering, Qingdao City University \\
     \textsuperscript{\rm 5}College of Computer Science and Technology, Ocean University of China \\
    jqleng24@m.fudan.edu.cn, kelvin.yakun.ju@gmail.com, fanhao@ouc.edu.cn
}

\usepackage{bibentry}
\begin{document}

\maketitle

\begin{abstract}
Surface-from-gradients (SfG) aims to recover a three-dimensional (3D) surface from its gradients. Traditional methods encounter significant challenges in achieving high accuracy and handling high-resolution inputs, particularly facing the complex nature of discontinuities and the inefficiencies associated with large-scale linear solvers. Although recent advances in deep learning, such as photometric stereo, have enhanced normal estimation accuracy, they do not fully address the intricacies of gradient-based surface reconstruction. To overcome these limitations, we propose a Fourier neural operator-based Numerical Integration Network (FNIN) within a two-stage optimization framework. In the first stage, our approach employs an iterative architecture for numerical integration, harnessing an advanced Fourier neural operator to approximate the solution operator in Fourier space. Additionally, a self-learning attention mechanism is incorporated to effectively detect and handle discontinuities. In the second stage, we refine the surface reconstruction by formulating a weighted least squares problem, addressing the identified discontinuities rationally. Extensive experiments demonstrate that our method achieves significant improvements in both accuracy and efficiency compared to current state-of-the-art solvers. This is particularly evident in handling high-resolution images with complex data, achieving errors of fewer than 0.1 mm on tested objects.
\end{abstract}


\section{Introduction}

\begin{figure}[t]	
\centering
\includegraphics[width=\linewidth]{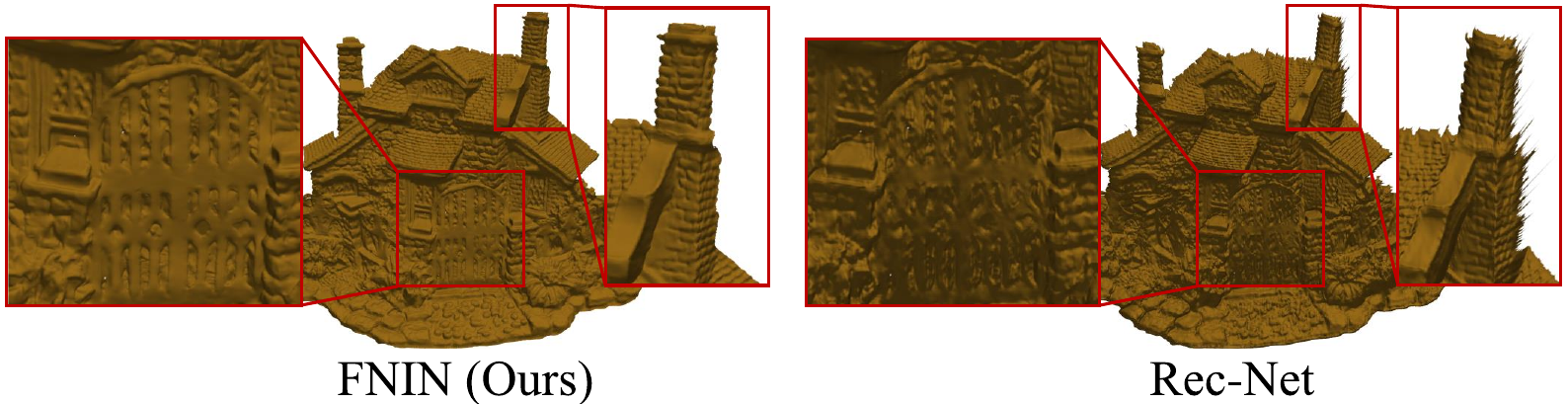}
	\caption{Reconstruction of the ``HOUSE'' in the LUCES \cite{luces} dataset. Previous Rec-Net \cite{lichy} introduces sharp features and distortion at discontinuities, while our FNIN preserves the desired details.}
	\label{illustration}  
\end{figure}

Numerical integration is foundational in computational mathematics, supporting various tasks in computer vision. Among these, reconstructing a 3D surface from its gradients, known as normal integration \cite{survey}, is crucial for accurate surface reconstruction. This process has become increasingly important as a follow-up step to photometric stereo (PS) \cite{shi2019benchmark,ju2023efficient,ju2024deep}, a method that estimates 3D surface normals from varying light reflectance. Recent advancements in PS have greatly improved normal estimation by overcoming the limitations of the Lambertian reflectance model \cite{psfcn,cnnps}. To achieve precise 3D mesh reconstructions, Surface-from-gradients (SfG), as shown in Fig. \ref{illustration}, emerges as a natural progression of these PS advancements. Furthermore, methods like normal integration are also essential in other vision tasks, such as computed tomography \cite{ct1} and image restoration \cite{ir1, yang2022sir}.

Although various strategies have been employed to tackle the SfG problem, existing methods often struggle with noise and integrability issues \cite{triple,survey,ipf}. Most advanced approaches are optimization-based and rely on hand-crafted features \cite{bini,ae}. However, these features often oversimplify real-world complexities. Poorly designed objectives can also lead to high computational costs or severe numerical instability, hindering practical application and complicating integration with other algorithms, like the PS normal estimation network in \cite{lichy}.

\begin{figure*}[!t]
	\centering
	\includegraphics[width=0.68
\linewidth]{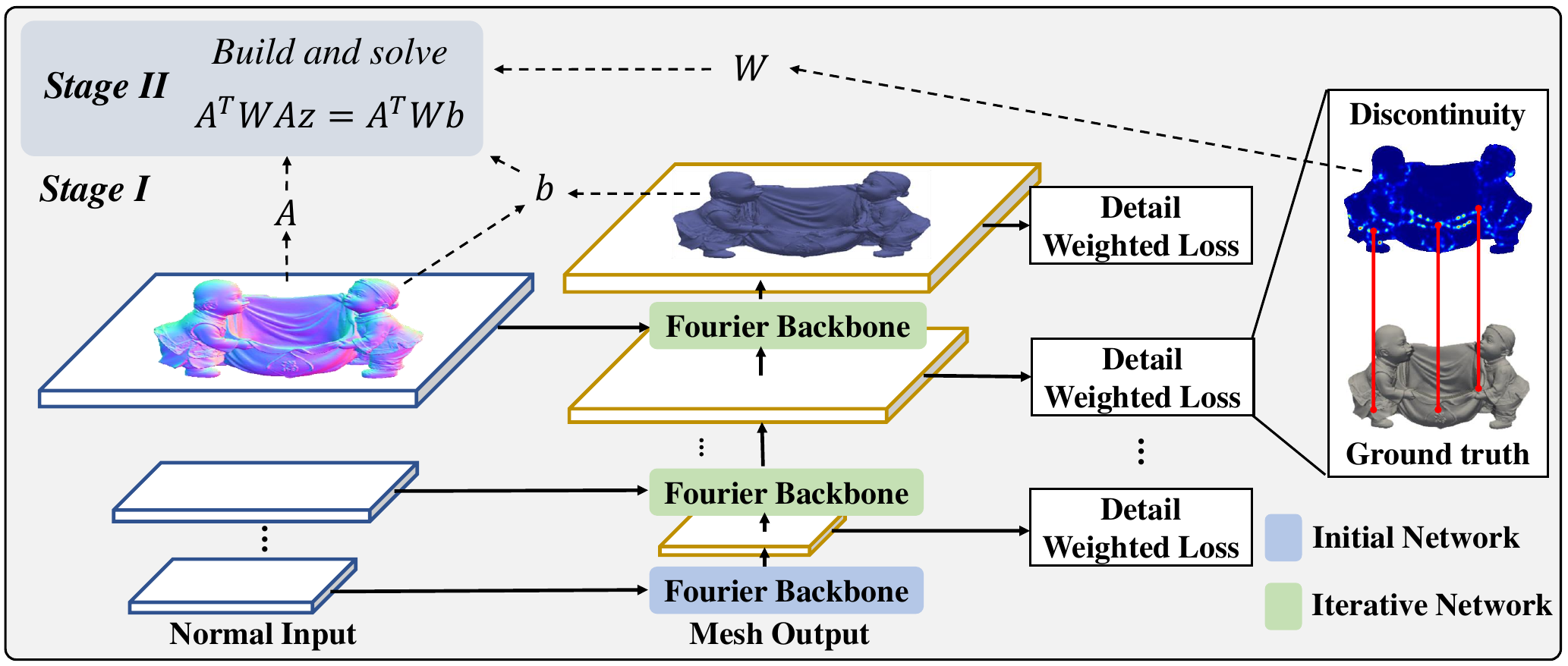}
	
	\caption{The Fourier neural operator-based numerical integration network operates within a two-stage framework. In stage I, the input normals are downsampled to different resolutions. In this iterative structure, the output depth is initially set by the initial network (blue) and subsequently refined by the iterative network (green). At each resolution, the Fourier backbone receives the upsampled output from the previous resolution and performs integration under the supervision of a detail weighted loss to preserve discontinuities. In stage II, the result is refined through a one-step weighted least squares optimization.}
	\label{scheme}  
\end{figure*}

Inspired by advancements in other vision tasks \cite{ct2, ir2, inverse, tvnet}, replacing optimization-based methods with end-to-end deep CNNs offers a promising improvement. However, straightforward applications \cite{lichy} have not achieved satisfactory results due to CNNs' reliance on discretization and inherent bias toward locality, which is problematic for the Surface-from-gradients (SfG) problem that depends on global information. Fortunately, a recently proposed neural network architecture called the neural operator (NO), originally developed for solving partial differential equations (PDEs) in computational physics \cite{deeponet, no, fno}, presents a potential solution. Neural operators leverage operator theory to learn mappings between infinite-dimensional function spaces without relying on PDEs as a physics prior. This capability enables us to introduce a learning-based approach to tackle the SfG problem effectively.

In this article, we propose a novel Fourier neural operator-based Numerical Integration Network (FNIN) to address the SfG problem using a learning-based approach. A key feature that distinguishes the SfG problem from typical NO equations is that, the discontinuities not satisfying the PDEs significantly influence the solution and tend to be extremely sparse, which will lead to confusion if learned pixel wise. To address this, we decouple the integration process of smooth area from optimization of discontinuity and introduce a two-stage scheme (Fig. \ref{scheme}). In the first stage, FNIN performs integration based on the formulated PDEs, approximating the solution through a Fourier neural operator (FNO) \cite{fno} in Fourier space, maintaining spatial globality. FNIN is constructed in a Rec-Net-like structure \cite{recnet} to ensure solution uniqueness. To detect discontinuities and mitigate disturbances, we incorporate an attention network, inspired by \cite{attention,ju2022normattention}, to generate an adaptive attention map. 
Consequently, we creatively obtain the solution at smooth areas without optimization and transferring the discontinuities into a relative weight.
% In the second stage, we address issues beyond the PDEs, primarily those from discontinuities. We formulate a weighted least squares problem following \cite{bini}, but perform the optimization only once, as discontinuities are detected in the first stage, making the process more time-efficient. 

In the second stage, we globally optimize the sparse discontinuities to avoid determining huge jumps pixel wise. We formulate a weighted least squares problem following \cite{bini}, but perform the optimization only once, as discontinuities are detected in the first stage, making the process more time-efficient. Our approach achieves state-of-the-art results on several objects from the DiLiGenT \cite{diligent} and LUCES \cite{luces} datasets, demonstrating the novelty of our method. Additionally, we conduct experiments to validate our approach further. Our method excels in handling high-resolution images with complex data, achieving errors of fewer than 0.1 mm on tested objects.

Our contributions are summarized as follows:
\begin{itemize}
\item We propose a Fourier neural operator-based integration network that performs integration on smooth areas and identifies discontinuities using a detail weighted loss.
\item We develop an optimization process that ensures optimal solutions at discontinuities through a weighted least squares equation.
\item Our experiments show that the proposed approach achieves exceptional accuracy and efficiency, making it a promising tool for future research.
\end{itemize}
\begin{links}
\link{Code}{https://github.com/nailwatts/FNIN}
\end{links}

\section{Related Work}
Our work is inspired by many existing studies but offers substantial novelty. In this section, we review several related works and highlight the key differences between them and our FNIN.

\subsection{SfG Methods Based on Optimization}

In the SfG problem, the ultimate goal is to solve the following unified partial differential equation
\begin{equation}
\scriptsize
\nabla z=\textbf{g},
\label{equation1}
\end{equation}
where $z$ represents the depth from the camera, and $\textbf{g}=[p,q]^T$ denotes the gradient map \cite{survey}. Ideally, when the surface is smooth, the solution should be independent of the integration path. However, in practice, due to natural discontinuities in real-world objects, perfect integrability is rarely achievable.

\begin{figure*}[!t]
\centering
\includegraphics[width=0.72\linewidth]{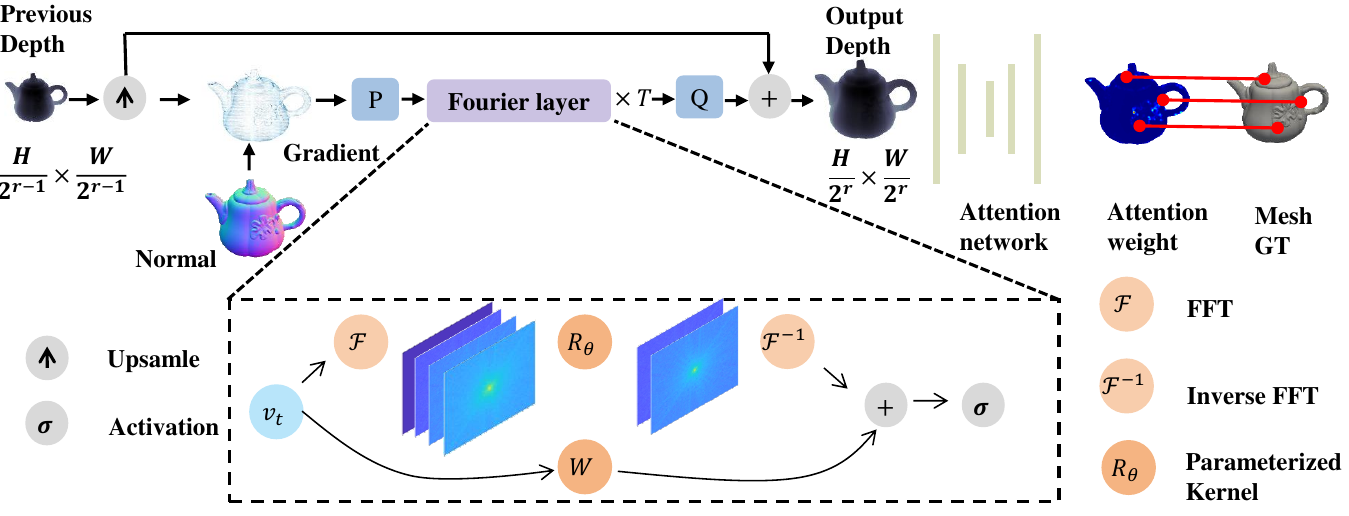}
	
	\caption{Integration is performed at an example resolution. Gradient \textbf{g} for the input is computed using the normal and the upsampled output from the previous resolution with Eq. \ref{equation2}. The integration is then carried out within NO framework, with the solution approximated in Fourier space. Finally, the relative weight for discontinuities is estimated by the attention network.}
	\label{forward}  
\end{figure*}

Recent works \cite{bini,ae} struggled with a coupled depth and discontinuity weight, requiring repeated optimization to account for discontinuities, which consumes significant time. In contrast, the proposed FNIN leverages a neural network that directly approximates the solution without relying on the PDEs, allowing us to decouple discontinuity detection from optimization. This data-driven detection method makes our approach highly innovative.

\subsection{PDE Solvers Based on Neural Operator}

The Neural Operator (NO) is a series of architectures designed to solve PDEs \cite{morton2005numerical} in a purely data-driven manner. The core idea is to convert PDEs into a mapping between the parameter space and the solution space. Consider a general form of PDEs defined on $D\subseteq \mathbb{R}^d$
\begin{equation}
\scriptsize
\begin{matrix}
(\mathcal{L}_{\alpha}z)(\textbf{u}) = f(\textbf{u}) \quad \textbf{u}\in D, \\
z(\textbf{u})=0 \quad \textbf{u}\in \partial D,
\end{matrix}
\label{no_equation}
\end{equation}
where $z:D\rightarrow\mathbb{R}$ represents the solution in the Banach space $\mathcal{Z}\subseteq\mathbb{R}^{d_z}$, and $a:D\rightarrow\mathbb{R}$  is the parameter in the Banach space $\mathcal{A}\subseteq\mathbb{R}^{d_a}$ that determines the operator $\mathcal{L}_{\alpha}$. The goal is to approximate the solution map $\mathcal{F}^{\dag}:\mathcal{A}\rightarrow\mathcal{Z}$ by training a neural network with finite-dimensional parameters $\theta^{\dag}\in\Theta$, such that $\mathcal{F}:\mathcal{A}\times \Theta\rightarrow\mathcal{Z}$ satisfies $\mathcal{F}(\cdot;\theta^{\dag})\approx\mathcal{F}^{\dag}$.

Li et al. \shortcite{no} observed that the operator learning task described above is equivalent to an empirical risk minimization problem, which can be solved through an iterative update. For $v_t\in\mathbb{R}^{d_v}, t=0,1,...,T$, the update is given by
\begin{equation}
\scriptsize
	v_{t+1}(\textbf{u})=\sigma(Wv_t(\textbf{u})+\mathcal{K}_{\theta}(a;\theta)v_t(\textbf{u})),
	\label{update}
\end{equation}
where $\sigma:\mathbb{R}\rightarrow\mathbb{R}$ is a non-linear activation function, $W:\mathbb{R}^{d_v}\rightarrow\mathbb{R}^{d_v}$ is a linear transformation, and the kernel integration is defined as
\begin{equation}
\scriptsize
	\mathcal{K}_{\theta}(a)(\textbf{u}):=\int_{D}\kappa_{\theta}(\textbf{u},\textbf{y},a(\textbf{u}),a(\textbf{y});\theta)v_t(\textbf{y})d\textbf{y},
	\label{no_kernel}
\end{equation}
where $\kappa_{\theta}:\mathbb{R}^{2(d+d_a)}\rightarrow\mathbb{R}^{d_v\times d_v}$ is a neural network parameterized by $\theta\in\Theta$. Typically, a NO first lifts the input $a\in \mathcal{A}$ into a higher-dimensional representation by $P:\mathbb{R}^{d_a}\rightarrow\mathbb{R}^{d_v}$,such that $v_0(\textbf{u})=P(a(\textbf{u}))$ and projects $v_T$ back by $z(\textbf{u})=Q(v_T(\textbf{u}))$, where $Q:\mathbb{R}^{d_v}\rightarrow\mathbb{R}^{d_z}$. 

In this work, the choice of NO architectures is flexible. We apply the popular FNO \cite{fno} based on our assumptions, but it should be noted that these components can be replaced with other NO architectures, such as \cite{Transformer, cno}, if necessary. However, the most significant difference is that, unlike most physical equations, our SfG problem is characterized by severe non-uniqueness and discontinuities, which heavily influence the solution. To address this, we construct a Rec-Net \cite{recnet} structure to maintain solution uniqueness and incorporate an attention network with a detail weighted loss, which helps suppress discontinuities during training. 

\section{The Proposed FNIN}

In this section, we introduce our Fourier neural operator-based numerical integration network (FNIN) with two stages. Fig. \ref{forward} illustrates the evaluation performed at one of the resolutions.

\subsection{Stage I of FNIN: Approximating in Fourier Space}

\subsubsection{Variants of SfG equation in a Neural Operator settings}
As Eqs. \ref{no_equation}-\ref{no_kernel} apply universally to the class of equations represented by Eq. \ref{no_equation}, a straightforward approach to solving Eq. \ref{equation1} is to use a mature NO architecture. However, previous research \cite{survey,dct} has shown that enforcing the Dirichlet boundary condition (setting the boundary value to 0) as implied by Eq. \ref{no_equation}, may lead to problematic solutions, despite ensuring uniqueness. To avoid this conflict with the natural boundary condition in Eq. \ref{equation1}, a proper transformation is needed.

Inspired by \cite{lichy}, we introduce an additional variable $\hat{z}$, which serves as an approximate solution to Eq. \ref{equation1}, such that
\begin{equation}
\scriptsize
	\begin{matrix}
		\nabla (z- \hat{z})=\textbf{g} - \nabla \hat{z} \quad \textbf{u}\in D, \\
		z- \hat{z}=0 \quad \textbf{u}\in \partial D,
	\end{matrix}
	\label{equation2}
\end{equation}
where $\textbf{g}=[p,q]^T$, and $\textbf{u}$ represents the image coordinates. Ideally, if $z$ and $\hat{z}$ are smooth and fully satisfy Eq. \ref{equation1}, Eq. \ref{equation2} would become an identity with a unique solution, $z\equiv\hat{z}$. In practice, we cannot provide such a solution, so we opt for a learning-based approach. Following \cite{lichy}, and considering the lack of detail in the synthetic dataset \cite{recnet}, we approximate $\hat{z}$ using the upsampled output from a lower resolution, and $\nabla \hat{z}$ is approximated by its central difference.

\subsubsection{Integration operator in spatial domain}
After transforming our SfG problem into an NO setting, the assumption of kernel integration in Eq. \ref{no_kernel} determines the specific architecture of the NO. It's necessary to guarantee that our assumption will not break the uniqueness of the solution.

\begin{figure}[t]
	\centering
	\includegraphics[width=\linewidth]{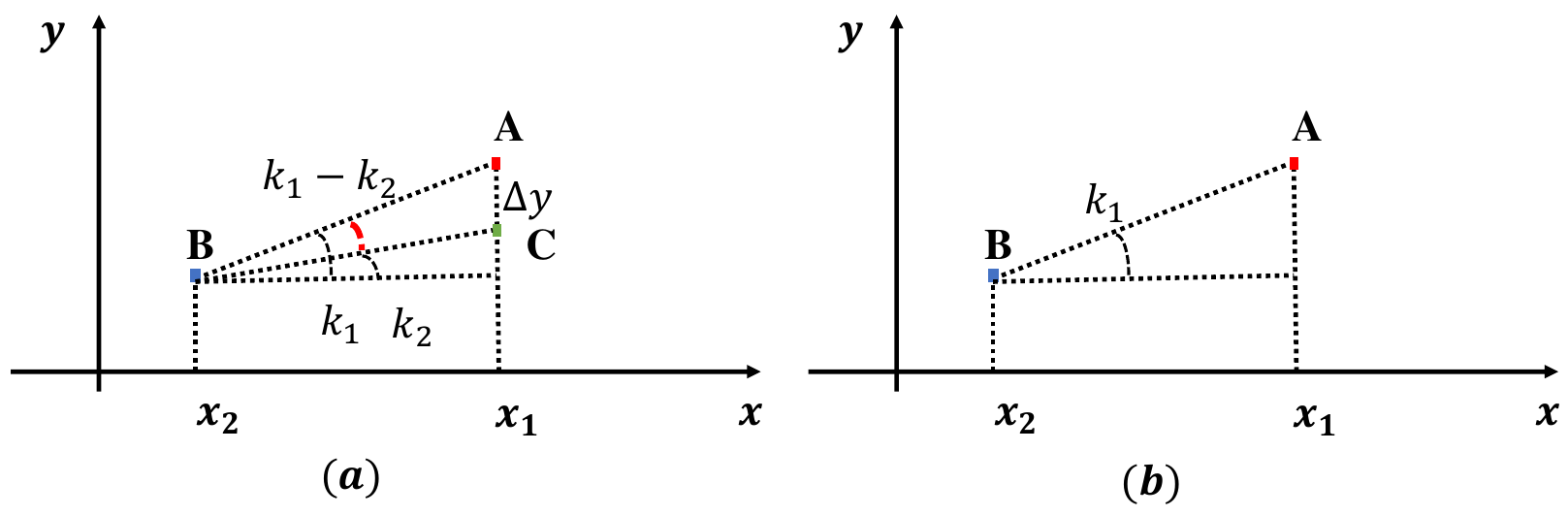}
	
	\caption{A simple example of (a) our assumption and (b) original natural boundary condition. Our assumption on given approximation green point C ensures the uniqueness otherwise the solution is ambiguous because  value at other coordinates (blue point B) is also unknown. }
	\label{example}
\end{figure}

As shown in Fig. \ref{example}, in a 1D scenario, the green point C represents our known approximated solution, and our goal is to derive the value at the red point A. Our assumption is that $\Delta y$ depends only on the distance $x_1-x_2$ between the red point A and the blue point B, and the difference in slope $k_1-k_2$. Here, $\textbf{g} - \nabla \hat{z}$ corresponds to $k_1-k_2$, and $\hat{z}$ corresponds to the green point C. Once we derive $\Delta y$, we can determine the desired $z$. This example also illustrates the non-uniqueness issue caused by the original boundary condition, as the value at A cannot be determined solely by $x_1-x_2$ and its slope $k_1$ as the value at B is also unknown.

Therefore, by applying our assumption to Eq. \ref{no_kernel}, we obtain
\begin{equation}
\scriptsize
	\mathcal{K}_{\theta}(\textbf{u}):=\int_{D}\kappa_{\theta}(\textbf{u}-\textbf{y};\theta)v_t(\textbf{y})d\textbf{y},
	\label{our_kernel}
\end{equation}
where $v_t(\textbf{y})$ corresponds to the lifted $\textbf{g} - \nabla \hat{z}$, and the dependence on the distance between coordinates is incorporated into the kernel. Clearly, this forms a perfect convolution with $\kappa_{\theta}$ as the kernel, enabling operator approximation in Fourier space. In fact, Eq. \ref{our_kernel} can be interpreted as adaptively determining the integration path. Compared to earlier works \cite{early, tree}, our parameterized kernel offers significantly greater robustness.

\subsubsection{Operator approximation in Fourier space}

As a convolution operator, a straightforward approach is to train a CNN to approximate Eq. \ref{our_kernel}. While increasing the receptive field could cover the entire domain $D$, the theory of effective receptive field \cite{erf} suggests that the network may struggle to focus globally, especially when locality is not included in Eq. \ref{our_kernel}. Moreover, since Eq. \ref{our_kernel} is meant to operate on a continuous domain rather than a discretized one, we choose to approximate it in Fourier space.

Consider the Fourier transform $\mathcal{F}$ of a function $f:D\rightarrow \mathcal{R}^{d_v}$ and its inverse $\mathcal{F}^{-1}$. By applying the convolution theorem, Eq. \ref{our_kernel} becomes
\begin{equation}
\scriptsize
	\mathcal{K}_{\theta}(\textbf{u})=\mathcal{F}^{-1}(R_{\theta}(\mathcal{F}v_t))(\textbf{u}),
\end{equation}
where $R_{\theta}$ is parameterization tensor. As we cannot approximate infinite terms of the Fourier series, we truncate it with a maximum node $k_{max}$. Further detailed discussion of FNO including the case under discrete Fourier transform can be referred to in \cite{fno}.

\subsubsection{Adaptive discontinuity detection} 
So far, we have considered integration in major smooth areas. In theory, our kernel in Eq. \ref{our_kernel} can handle jumps at discontinuities. However, since discontinuities are much sparser than smooth areas, the network may make incorrect jumps, disrupting the approximation in other regions. To address this issue inspired by other vision tasks \cite{fang2024your}, we transfer them into a relative weight depending on their extension to be suppressed.

During the training process, we adopt a self-learning attention map. Unlike hand-crafted features in previous methods, we employ an extractor and regressor for attention, which can be formulated as follows
\begin{equation}
\scriptsize
	\begin{matrix}
		\Phi = f_{\theta_{ae}}(\Delta z;\theta_{ae}),\\
		\omega = \mathrm{normalize}( f_{\theta_{ar}}(\Phi;\theta_{ar})),
	\end{matrix}
\end{equation}
where $f_{\theta_{ae}}$ and $f_{\theta_{ar}}$ are both three-layer CNNs, $z$ represents the current depth estimation, and $\Delta$ represents the discontinuity modeled by the difference between the one-sided limit and its value measured in point-to-plane distance \cite{ipf}, following \cite{bini,ae}. The variable $\omega$ is the relative discontinuity weight, which is normalized to maintain values within $[0,1]$.

To enable the attention network to detect discontinuities, we designed a detail weighted loss inspired by the ideas in \cite{attention,ju2022normattention}, which has been used for detecting discontinuities in normals. For each resolution $r$, the normalized attention map $\omega$ balances the trade-off between the normal calculated from adjacent pixels and the absolute depth value. The loss function is expressed as
\begin{equation}
\scriptsize
	\mathcal{L}_r=\frac{1}{N}\sum_{i=1}^N((1-\omega_i)||z_i-z^{gt}_i||_1+\gamma \omega_i||\textbf{n}_i(\textbf{x}) - \textbf{n}_i(\textbf{x}^{gt})||_1),
	\label{attention_weighted}
\end{equation}
where $N=|D|$ represents the number of all valid points, $\omega_i$ is the weight at pixel $i$ in the attention map $\omega$, and $\gamma$ is a hyperparameter set to 0.25. The terms $z_i$ and $z^{gt}_i$ are the estimated and ground truth depths at pixel $i$, respectively, while $\textbf{x}$ and $\textbf{x}^{gt}$ represent the point clouds calculated via camera projection. The normals $\textbf{n}_i(\textbf{x})$ and $\textbf{n}_i(\textbf{x}^{gt})$ are computed using point clouds with central differences. Although many previous works have used similar losses to enhance high-frequency areas, we use the detail weighted loss to suppress discontinuities, which is a completely different approach. Combining all resolutions, the network is supervised by
\begin{equation}
\scriptsize
	\mathcal{L} = \sum_{r=1}^R \mathcal{L}_r
\end{equation}
where $R$ is the number of different resolutions.

\begin{table*}[t]
 \centering
 \scriptsize
	\setlength{\tabcolsep}{1.5mm}{

		\begin{tabular}{c|ccccccccccccccc|c}
			\toprule
			Assumption & Method &Ball&Bell&Bowl&Buddha&Bunny&Cup&Die& Hippo& House&Jar&Owl&Queen&Squirrel&Tool&Runtime (s) \\
			\midrule

		&HB&2.12&3.77&4.09&\ding{56}& 5.80&1.25&3.30&5.95 &\textbf{8.61} &7.46 &5.43 &6.68 &2.22 & 5.72&$>6000$\\

			&FFT&0.50&4.29&4.66&23.78& 4.85&0.80&2.28&5.59&9.57&3.57&4.77&5.79&1.64&4.96&0.23\\

			&DCT&0.30&1.45&2.13&22.56& 4.07&0.81&1.27&4.67&9.08&2.01&4.18&6.01&1.54&4.71&0.49\\
			
			Smooth&DF&0.46&2.54&\ding{56}&\ding{56}&\ding{56}&0.37&\ding{56}&\ding{56}&\ding{56}& \ding{56} & 4.15 &\ding{56}&\ding{56} & 2.71 & 136.75 \\
			
			&DP&0.17&0.33&0.38&\ding{56}&2.66&0.09&0.66&1.77&9.30& 1.12 & 3.62 & 4.09& \textbf{1.11} & 1.88 & 6.64 \\
			
			&ZS & \multicolumn{15}{c}{OOM} \\
			
			&IPF & \ding{56} & 0.98 & 1.15 & 6.93 & 3.09 & \ding{56} & \ding{56} & 1.68 & \ding{56} & 3.06 & 3.86 & 1.86 & 1.84 & 1.11& 45.22 \\

			\midrule
			
			& $L^1$ &0.29&1.47&2.12&6.88& 3.81&0.21&1.21&4.61&9.03&1.96&2.89&4.42&1.51&4.42&110.77\\
			
			Discontinue&AD&0.18&0.36&0.46&3.40& 2.62&0.11&0.68&2.06&9.26&1.41&3.22&3.71&1.15&0.80&59.70\\

			&BiNI$^\star$&0.54&0.96&1.16&3.59& 2.92&0.70&2.03&1.32&11.13&3.20&\textbf{1.62}&\textbf{0.90}&1.43&1.11&7.31\\
			
			\midrule
			
			& Rec-Net$^\star$  & 0.93 & 1.38 & 1.28 & 18.44 & 2.25 & 1.63 & 1.07 & 1.79 & 15.60 & 2.05 & 1.72 & 2.52 & 2.48 & 4.16 & 0.09\\
			
			Learning& FNIN$^\star$ & \textbf{0.15} & \textbf{0.12} & 0.08 & \textbf{3.35} & 2.45 & \boldmath{$<0.01$}& 0.70& 1.26 & 9.96& 0.28 & 3.77 & 3.38 & 1.18 & 0.28 & 1.81\\
			& FNIN-S$^\star$ & \textbf{0.15} & 0.15 & \textbf{0.05} & 3.63 & \textbf{2.37} & 0.03& \textbf{0.59} & \textbf{1.21} & 10.50 & \textbf{0.27} & 3.39 & 2.63 &\textbf{ 1.11} &\textbf{0.25} & 1.28\\
			\bottomrule 
		\end{tabular}
	}
    \caption{Results on the LUCES \cite{luces} Dataset. The metric for each object is Mean Absolute Error (MAE) in millimeters (mm). \ding{56} indicates a numerical failure, and OOM indicates an out-of-memory error. Methods marked with $\star$ are GPU-compatible and were tested on the same GPU. Best result for each object is highlighted in bold.\label{luces_tab}}
\end{table*}

\begin{table}[t]
	\scriptsize
	\centering
	\setlength{\tabcolsep}{0.75mm}{
		\begin{tabular}{cccccccccc}
			\toprule
			Method &Bear&Buddha&Cat&Cow&Goblet&Harvest&Pot 1& Pot 2& Reading \\
			\midrule
			FFT&3.43&5.51&7.23&2.89&15.69&3.37&2.54&1.72&6.13\\			
			DST&3.41&5.71&7.26&2.89&15.66&3.39&2.56&1.73&6.16 \\
			DCT&2.99&5.72&7.29&2.80&15.26&3.34&2.48&1.69&5.75\\

			ZS&1.82&3.80&1.72&1.16&12.72&11.17&2.00&0.78&7.09 \\
			
			IPF&0.89&\textbf{3.33}&1.48&0.86&11.23&11.11&1.73&0.64&6.56 \\
			
			\midrule
			
			WLS&2.52&3.50&6.12&2.10&14.75&\textbf{2.60}&8.00&1.15&4.54\\
                TV & 4.30 & 4.72 & 3.71 & 2.00 &\textbf{9.89} & 14.47 & 3.38 & 1.35 & \textbf{3.30} \\

                MS  &\multicolumn{9}{c}{\ding{56}}\\

			\midrule
			
			 Rec-Net  &\multicolumn{9}{c}{\ding{56}}\\
			
			FNIN & 1.35 & 3.74 & 1.69 & 1.18 & 12.61 & 11.17 & 1.96 & 0.82 &7.07 \\ 
			 FNIN-S & \textbf{0.11} & 3.52 & \textbf{1.41} & \textbf{0.66} & 11.53 & 12.42 & \textbf{1.53} & \textbf{0.49} & 6.38  \\
			
			\bottomrule
		\end{tabular}
	}
    \caption{Results on the DiLiGenT \cite{diligent} Dataset. Best result for each object is highlighted in bold. \label{diligent_tab}}
\end{table}

\begin{figure}[t]	
\centering
\includegraphics[width=0.8\linewidth]{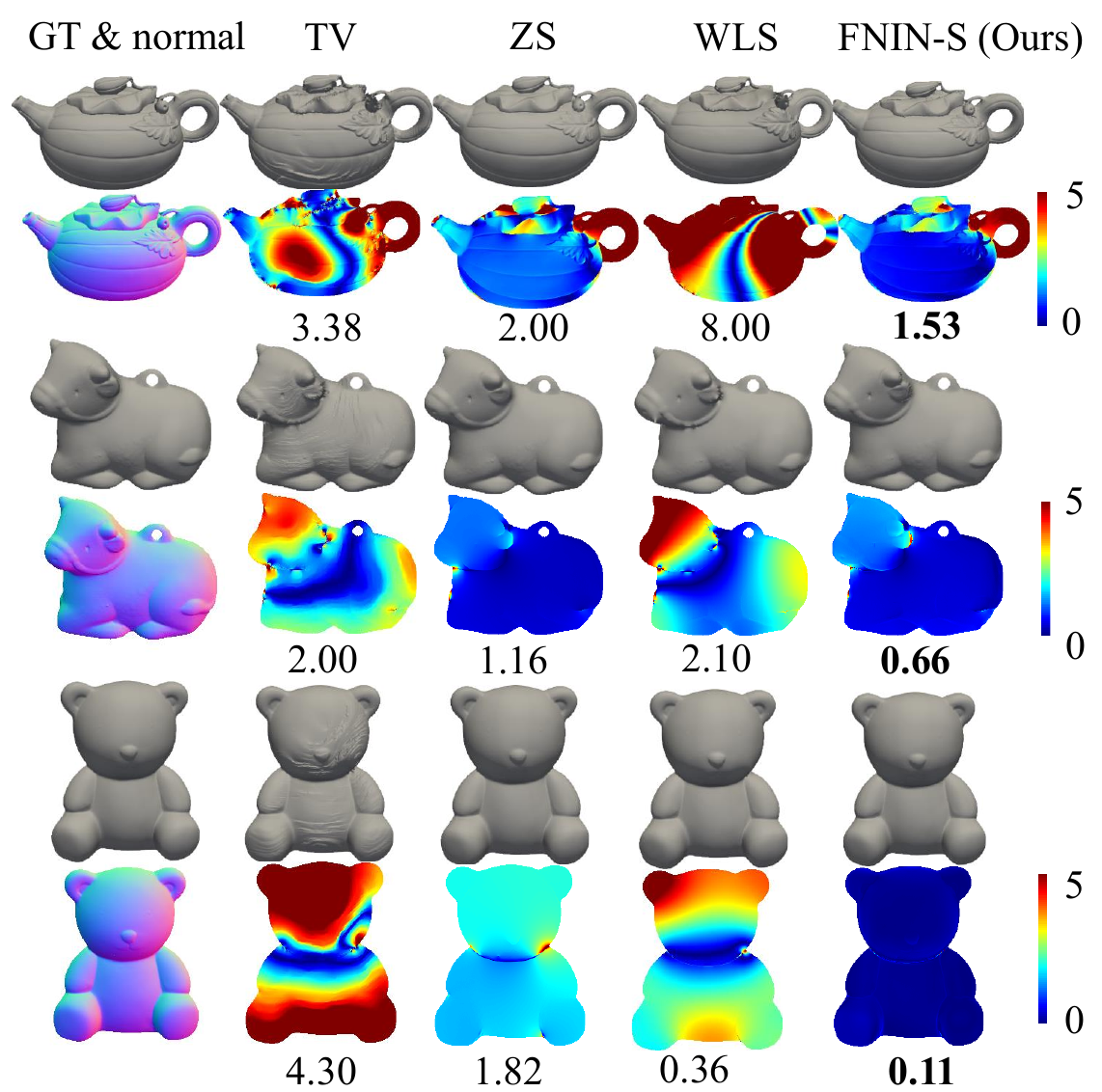}
	\caption{Qualitative results on ``\textsc{BRAR}'', ``\textsc{COW}'' and ``\textsc{POT 1}'' objects from DiLiGenT \cite{diligent} Dataset. The black numbers under the error maps indicate the MAE (mm). The best result for each object is highlighted in bold.}
	\label{diligent}  
\end{figure}

\begin{figure*}[t]
\centering
\includegraphics[width=0.7\linewidth]{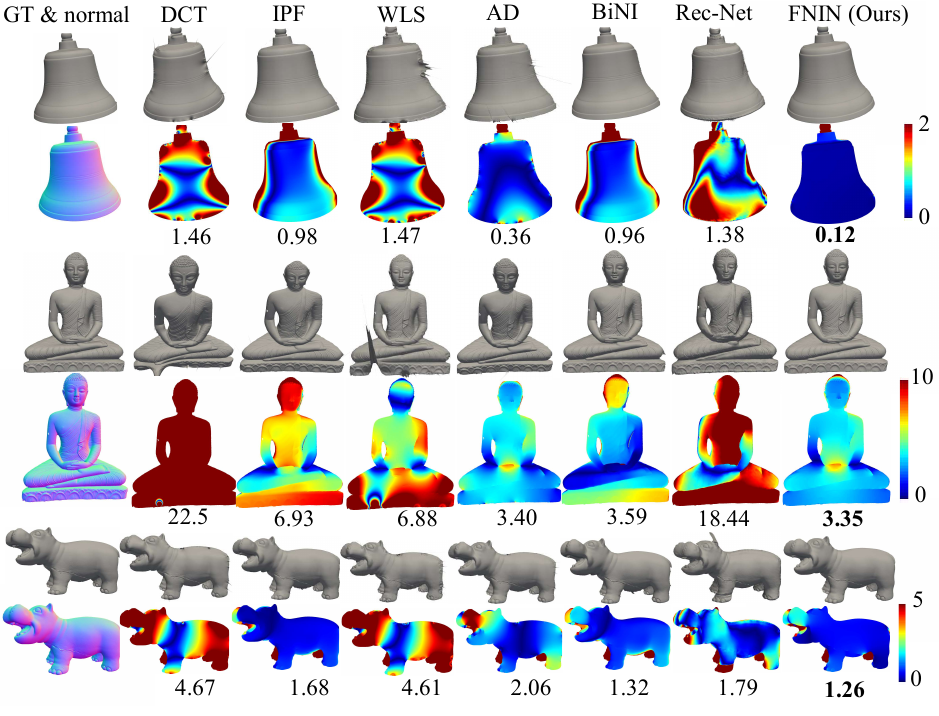}
	
	\caption{Qualitative results on ``\textsc{BELL}'', ``\textsc{BUDDHA}'' and ``\textsc{HIPPO}'' objects from LUCES \cite{luces} Dataset. Ground truth, input normals, estimated meshes, and absolute difference maps are shown in each pair of rows. The black numbers under the error maps indicate the MAE (mm). The best result for each object is highlighted in bold.}
	\label{luces}  
\end{figure*}

\subsection{Stage II of FNIN: Optimizing Discontinuity }

To this end, our integration network has performed integration on most smooth areas and provided a discontinuity map. The remaining issue is to determine the jump at discontinuities through optimization.

The key idea is to suppress the effect of discontinuities in the objective function using weights, where the attention mechanism plays a crucial role. Since our attention-based discontinuity map is point-wise, meaning it assigns a value to each pixel, we can formulate a convex weighted least squares problem with four directions in the discrete case by
\begin{equation}
\scriptsize
	\begin{aligned}
		\mathop{\min}_{z(u,v)} \sum_{D} &\frac{w_r}{2}(n_3\partial_u^+z+n_1)^2+\frac{w_l}{2}(n_3\partial_u^-z+n_1)^2\\
		+&\frac{w_t}{2}(n_3\partial_v^+z+n_2)^2+\frac{w_b}{2}(n_3\partial_v^-z+n_2)^2\\
		+&\lambda(z(u,v)-z_R(u,v))^2,
	\end{aligned}
	\label{our_objective}
\end{equation}
where $w_r, w_l, w_t, w_b$ are the weights at adjacent pixels in $\omega$ from the right, left, top, and bottom directions, respectively.  $z_R$ is the log depth output from FNIN, and $\lambda=1e^{-3}$ is a fixed constant. We directly solve the normal equation of Eq. \ref{our_objective}, similar to the approach in \cite{bini}. Consider the flattened vectors $\mathbf{n}_1, \mathbf{n}_2, \mathbf{n}_3$, and $\mathbf{z}_R \in \mathbb{R}^N$, where the depth map $\mathbf{z} \in \mathbb{R}^N$, flattened in the same order, is given by
\begin{equation}
\scriptsize
\textbf{A}^T\textbf{W}\textbf{A}\textbf{z}=\textbf{A}^T\textbf{W}\textbf{b},
	\label{matrix}
\end{equation}
where
	\begin{equation}
		\scriptsize
		\textbf{A}=\begin{bmatrix}
			\textbf{N}_z\textbf{D}_u^+ + \lambda\boldsymbol{1} \\
			\textbf{N}_z\textbf{D}_u^- +\lambda\boldsymbol{1}\\
			\textbf{N}_z\textbf{D}_v^+ +\lambda\boldsymbol{1}\\
			\textbf{N}_z\textbf{D}_v^- +\lambda\boldsymbol{1}
		\end{bmatrix}, \, \textbf{b}=\begin{bmatrix}
			-\textbf{n}_1 + \lambda\mathbf{z}_R \\
			-\textbf{n}_1 + \lambda\mathbf{z}_R \\
			-\textbf{n}_2 + \lambda\mathbf{z}_R \\
			-\textbf{n}_2 + \lambda\mathbf{z}_R
		\end{bmatrix}, \, \textbf{W} = diag\Bigg (\begin{bmatrix}
			\textbf{w}_r/2\\
			\textbf{w}_l/2 \\
			\textbf{w}_t/2 \\
			\textbf{w}_b/2
		\end{bmatrix}\Bigg),
	\end{equation}
and $\textbf{N}_z=diag(\mathbf{n}_3)$, $\textbf{D}_u^+, \textbf{D}_u^-, \textbf{D}_v^+, \textbf{D}_v^-\in \mathbb{R}^{N\times N}$ are four discrete partial derivative matrices, $\boldsymbol{1}\in \mathbb{R}^N$ is all-one vector. We ultimately obtain the desired depth map after solving Eq. \ref{matrix}.

\section{Experiments}

 Our FNIN is implemented in PyTorch \cite{pytorch}, with Eq.\ref{matrix} solved using the conjugate gradient method \cite{cg}. 
We used a synthetic dataset rendered by \cite{lichy}, consisting of 14 objects from the statue dataset \cite{statue}, with 1.75k training samples and 175 validation samples, each sized at $512\times 512$. To reduce memory consumption from large kernels, we applied Tucker factorization as in \cite{tfno}. For discontinuity detection, we additionally provide a sigmoid function (FNIN-S) \cite{bini}, handling discontinuities with varying intensities. 

 \begin{table*}[!t]
	\centering
 \scriptsize
	\setlength{\tabcolsep}{1.75mm}{
		\begin{tabular}{cccccccc}
			\toprule
			\multirow{2}{*}{Experiment} & \multirow{2}{*}{Method} & \multirow{2}{*}{Parameters} &  \multicolumn{2}{c}{\underline{DiLiGenT ($612\times 512$)}}  & \multicolumn{3}{c}{\underline{LUCES ($2048\times 1536$)}}  \\
			
			& & & MAE (mm) & RAM+VRAM (GB) &MAE (mm) & Runtime (s) & RAM+VRAM (GB)\\
			\midrule
			\multirow{2}{*}{FNIN}& w/o FNIN (BiNI, k=0) & -- & 4.59 & 1.6 + 1.0 & 2.79  & 0.91 & 2.0 + 2.0 \\
			&FNIN-S & 1.53M & 4.23 & 4.0 + 4.8 & 1.88 & 1.28 & 4.5 + 13.9 \\

			\midrule
			\multirow{2}{*}{Structure$^*$}& w/o attention & 1.05M & 5.97  & 3.7 + 4.5 & 4.02 & 0.21 & 3.9 + 12.5 \\
			& w/ mask concatenation & 1.53M & 15.15 & 3.7 + 4.49 & 3.96 & 0.50  & 4.1 + 13.5 \\
			\midrule
			\multirow{2}{*}{Discontinuity} & w/o discontinuity optimization & 1.53M & 5.91 & 3.7 + 4.8 & 3.00 & 0.39 & 4.1 + 13.5\\
			& FNIN & 1.53M & 4.55 & 4.1 + 4.8 & 1.92 & 1.66  & 4.6 + 13.5\\
			
			\bottomrule
		\end{tabular}
	}
    \caption{Ablation study of our approach. Discontinuity optimization is omitted in ``Structure$^*$''. ``Runtime'' shows the average evaluation time across all objects, and ``RAM+VRAM'' records the peak memory usage. \label{ablation}}
\end{table*}

\subsection{Comparison with Previous Methods}

 Previous optimization-based methods excel in generalization. Classic neural networks, based on the universal approximation theorem, often struggle with generalization errors \cite{deeponet}, making them less competitive with traditional numerical PDE solvers, especially for purely geometric problems like SfG. In contrast, our FNIN, derived from minimizing cost functionals in infinite dimensions, achieves exceptional generalization. We conducted a comparison where our learning-based method, without fine-tuning, was directly compared with state-of-the-art numerical methods. We used the DiLiGenT \cite{diligent} and LUCES \cite{luces} datasets. The DiLiGenT dataset contains 10 real-world objects scanned at $612\times 512$ resolution, while the LUCES dataset includes 14 objects at $2048\times 1536$ resolution, closer to common camera resolutions. This higher resolution challenges efficiency and detail preservation. We report the Mean Absolute Error (MAE) for each object, with results presented in Tables \ref{luces_tab} and \ref{diligent_tab} and visualized in Figs. \ref{diligent} and \ref{luces}.

\subsubsection{Comparison with methods based on smooth surface assumption}

For methods based on the smooth surface assumption, we selected several approaches: the symbolic Horn and Brooks (HB) method \cite{hb}, the fast Fourier transform (FFT) integrability enforcing method \cite{fft}, the discrete cosine transform (DCT), discrete sine transform (DST) integrability enforcing method \cite{dct}, the discretized functional (DF) variational method \cite{df}, the 2D Savitzky-Golay filters variational method (ZS) \cite{zs}, and the four-point inverse plane fitting (IPF) method \cite{ipf}.

Our approach outperforms others in both accuracy and stability. The attention network effectively prevents FNIN from being affected by large discontinuities (e.g., ``BUDDHA'' in Fig.\ref{luces}), and the discontinuity optimization step preserves these details, significantly improving mesh accuracy. Additionally, our iterative structure maintains numerical stability, minimizing spikes caused by exponential functions.

\subsubsection{Comparison with methods focusing on discontinuity preservation}

For methods focusing on discontinuity preservation, we selected the variational method using Split-Bregman iterations to solve the weighted least squares problem (WLS) \cite{wls}, the variational method using anisotropic diffusion (AD), isotropic total variation (TV),  Mumford and Shah Functional (MS) \cite{variational}, and the variational method using bilaterally weighted functional \cite{bini}.

Our approach outperforms these methods in capturing fine details. For objects like "BELL" in Fig.\ref{luces} and "BEAR" in Fig.\ref{diligent}, where discontinuities are small and difficult to detect, our method performs significantly better. Notably, our accuracy on the "CUP" object in Table \ref{luces_tab} is impressive, with an MAE below 0.01 mm. Additionally, FNIN integrates easily into other deep learning methods due to its efficiency on high-resolution images and ability to parallelize operations.

\subsubsection{Comparison with Rec-Net}

To our knowledge, Rec-Net \cite{lichy} is the only deep learning method designed for the SfG problem. Key differences between our approach and Rec-Net are: First, while both use iterative structures, Rec-Net cannot be formulated into Eq. \ref{our_kernel} due to CNNs' inherent locality, which prevents it from guaranteeing the uniqueness of Eq. \ref{equation2}. Second, Rec-Net fails to account for discontinuities, causing irregular jumps and hindering convergence. Third, simply concatenating the input and mask in Rec-Net can cause confusion during transfer, leading to numerical failures (Table \ref{diligent_tab}). Our FNIN addresses these issues, often outperforming traditional solvers, as shown in Table \ref{luces_tab}.

\subsection{Ablation Study}

In this section, we focus on the most crucial characteristics of our approach, including the effects of FNIN, the iterative structure, the attention network, and discontinuity optimization. Additionally, we analyze the computational cost and the handling of non-rectangular domains. 

\begin{figure}[t]
	\centering
	\includegraphics[width=\linewidth]{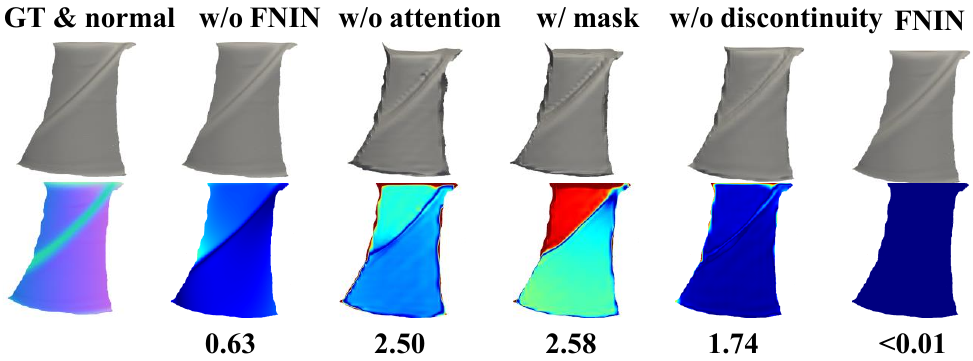}
	\caption{Ablation study on the ``\textsc{CUP}'' object from the LUCES \cite{luces} Dataset. The black numbers under the error maps indicate the MAE (mm).}
	\label{ablation_fig}  
\end{figure}

\subsubsection{Ablation on the effect of FNIN} Without FNIN, our approach degenerates into a smoother version of BiNI \cite{bini}, resulting in blurred discontinuities and reduced accuracy (Fig. \ref{ablation_fig}). Although optimization is performed only once, most runtime is spent solving the WLS problem, ensuring efficient training. The factorization we apply also facilitates integration into other deep learning algorithms.

\subsubsection{Ablation on the effect of attention} The attention mechanism in FNIN and the weights in discontinuity optimization suppress discontinuities at different levels. Attention in the detail-weighted loss prevents incorrect pixel-level normals from adjacent pixels. Without it, accuracy drops significantly at edges and discontinuities (Fig. \ref{ablation_fig}), as the PDEs in Eq. \ref{equation1} and Eq. \ref{equation2} do not account for discontinuities, necessitating the decoupling of these processes. The WLS weights balance the impact of discontinuities and determine their jumps. Both components are essential.

\subsubsection{Ablation on handling non-rectangular domain} Handling arbitrary domains is crucial, as shown by Rec-Net's instability on DiLiGenT. While the domain should ideally be embedded in the input \cite{dafno}, our attention network allows us to directly multiply the gradient by the mask, ensuring edge discontinuities. Even if FNIN is affected by boundaries, the attention mechanism and discontinuity optimization resolve the issues seamlessly, as demonstrated by FNIN and FNIN-S results.

\subsubsection{Ablation on discontinuity optimization} Finally, our network, designed to avoid being misled by jumps at sparse discontinuities, cannot recover them individually. Due to this conflict, pixel-level accuracy in regions of discontinuity is not advisable. Removing discontinuity optimization, as shown in Table \ref{ablation}, results in a significant drop in accuracy.

\section{Conclusion}

In this work, we developed a numerical integration neural network within a two-stage framework for the Surface-from-gradients (SfG) problem. In the first stage, the solution is approximated by a neural operator in Fourier space, using a detail-weighted loss to address discontinuities. In the second stage, the identified discontinuities are optimized using a weighted least squares equation to determine their jumps rationally. Our experiments demonstrate that our approach achieves state-of-the-art results with exceptional efficiency. 

\section*{Acknowledgements} 

This work was supported in part by the National Natural Science Foundation of China under Grant  42106193 and Grant 41927805.

\bibliography{aaai25}

\begin{thebibliography}{59}
\providecommand{\natexlab}[1]{#1}

\bibitem[{Badri, Yahia, and Aboutajdine(2014)}]{triple}
Badri, H.; Yahia, H.; and Aboutajdine, D. 2014.
\newblock Robust Surface Reconstruction via Triple Sparsity.
\newblock In \emph{Proceedings of the IEEE/CVF Conference on Computer Vision and Pattern Recognition}, 2291--2298.

\bibitem[{Bradshaw et~al.(2013)Bradshaw, Taubman, Todd, Magnussen, and Halmagyi}]{ct1}
Bradshaw, A.~P.; Taubman, D.~S.; Todd, M.~J.; Magnussen, J.~S.; and Halmagyi, G.~M. 2013.
\newblock Augmented Active Surface Model for the Recovery of Small Structures in CT.
\newblock \emph{IEEE Transactions on Image Processing}, 22(11): 4394--4406.

\bibitem[{Cao(2021)}]{Transformer}
Cao, S. 2021.
\newblock Choose a Transformer: Fourier or Galerkin.
\newblock In \emph{Proceedings of the Advances in Neural Information Processing Systems}, volume~34, 24924--24940. Curran Associates, Inc.

\bibitem[{Cao et~al.(2022)Cao, Santo, Shi, Okura, and Matsushita}]{bini}
Cao, X.; Santo, H.; Shi, B.; Okura, F.; and Matsushita, Y. 2022.
\newblock Bilateral Normal Integration.
\newblock In \emph{Proceedings of the European Conference on Computer Vision}.

\bibitem[{Cao et~al.(2021)Cao, Shi, Okura, and Matsushita}]{ipf}
Cao, X.; Shi, B.; Okura, F.; and Matsushita, Y. 2021.
\newblock Normal Integration via Inverse Plane Fitting with Minimum Point-to-Plane Distance.
\newblock In \emph{Proceedings of the IEEE/CVF Conference on Computer Vision and Pattern Recognition}, 2382--2391.

\bibitem[{Chen, Ng, and Zhao(2015)}]{ir1}
Chen, C.; Ng, M.~K.; and Zhao, X.-L. 2015.
\newblock Alternating Direction Method of Multipliers for Nonlinear Image Restoration Problems.
\newblock \emph{IEEE Transactions on Image Processing}, 24(1): 33--43.

\bibitem[{Chen, Han, and Wong(2018)}]{psfcn}
Chen, G.; Han, K.; and Wong, K.-Y.~K. 2018.
\newblock PS-FCN: A Flexible Learning Framework for Photometric Stereo.
\newblock In \emph{Proceedings of the European Conference on Computer Vision}, 3--19.

\bibitem[{Fan et~al.(2018)Fan, Huang, Gan, Ermon, Gong, and Huang}]{tvnet}
Fan, L.; Huang, W.; Gan, C.; Ermon, S.; Gong, B.; and Huang, J. 2018.
\newblock End-to-End Learning of Motion Representation for Video Understanding.
\newblock In \emph{Proceedings of the IEEE/CVF Conference on Computer Vision and Pattern Recognition}, 6016--6025.

\bibitem[{Fang et~al.(2024)Fang, Easwaran, Genest, and Suganthan}]{fang2024your}
Fang, X.; Easwaran, A.; Genest, B.; and Suganthan, P.~N. 2024.
\newblock Your Data Is Not Perfect: Towards Cross-Domain Out-of-Distribution Detection in Class-Imbalanced Data.
\newblock \emph{ESWA}.

\bibitem[{Frankot and Chellappa(1988)}]{fft}
Frankot, R.; and Chellappa, R. 1988.
\newblock A Method for Enforcing Integrability in Shape from Shading Algorithms.
\newblock \emph{IEEE Transactions on Pattern Analysis and Machine Intelligence}, 10(4): 439--451.

\bibitem[{Harker and O’Leary(2014)}]{df}
Harker, M.; and O’Leary, P. 2014.
\newblock {Regularized Reconstruction of a Surface from its Measured Gradient Field}.
\newblock \emph{Journal of Mathematical Imaging and Vision}, 51(1): 46--70.

\bibitem[{Hestenes and Stiefel(1952)}]{cg}
Hestenes, M.~R.; and Stiefel, E. 1952.
\newblock Methods of Conjugate Gradients for Solving Linear Systems.
\newblock \emph{Journal of research of the National Bureau of Standards}, 49: 409--435.

\bibitem[{Horn and Brooks(1986)}]{hb}
Horn, B.~K.; and Brooks, M.~J. 1986.
\newblock The Variational Approach to Shape from Shading.
\newblock \emph{Computer Vision, Graphics, and Image Processing}, 33(2): 174--208.

\bibitem[{Hsieh et~al.(1994)Hsieh, Liao, Ko, and Fan}]{wavelet}
Hsieh, J.-W.; Liao, H.-Y.; Ko, M.-T.; and Fan, K.-C. 1994.
\newblock Wavelet-based Shape from Shading.
\newblock In \emph{Proceedings of International Conference on Image Processing}, volume~2, 125--129.

\bibitem[{Ikehata(2018)}]{cnnps}
Ikehata, S. 2018.
\newblock {CNN-PS: CNN-Based Photometric Stereo for General Non-convex Surfaces}.
\newblock In \emph{Proceedings of the European Conference on Computer Vision}, 3--19.

\bibitem[{Jin et~al.(2017)Jin, McCann, Froustey, and Unser}]{inverse}
Jin, K.~H.; McCann, M.~T.; Froustey, E.; and Unser, M. 2017.
\newblock Deep Convolutional Neural Network for Inverse Problems in Imaging.
\newblock \emph{IEEE Transactions on Image Processing}, 26(9): 4509--4522.

\bibitem[{Ju et~al.(2020)Ju, Lam, Chen, Qi, and Dong}]{attention}
Ju, Y.; Lam, K.-M.; Chen, Y.; Qi, L.; and Dong, J. 2020.
\newblock Pay Attention to Devils: A Photometric Stereo Network for Better Details.
\newblock In \emph{Proceedings of the International Joint Conference on Artificial Intelligence}, 694--700.

\bibitem[{Ju et~al.(2023)Ju, Lam, Xiao, Zhang, Yang, and Dong}]{ju2023efficient}
Ju, Y.; Lam, K.-M.; Xiao, J.; Zhang, C.; Yang, C.; and Dong, J. 2023.
\newblock Efficient feature fusion for learning-based photometric stereo.
\newblock In \emph{ICASSP 2023-2023 IEEE International Conference on Acoustics, Speech and Signal Processing (ICASSP)}, 1--5. IEEE.

\bibitem[{Ju et~al.(2024)Ju, Lam, Xie, Zhou, Dong, and Shi}]{ju2024deep}
Ju, Y.; Lam, K.-M.; Xie, W.; Zhou, H.; Dong, J.; and Shi, B. 2024.
\newblock Deep Learning Methods for Calibrated Photometric Stereo and Beyond.
\newblock \emph{IEEE Transactions on Pattern Analysis and Machine Intelligence}.

\bibitem[{Ju et~al.(2022)Ju, Shi, Jian, Qi, Dong, and Lam}]{ju2022normattention}
Ju, Y.; Shi, B.; Jian, M.; Qi, L.; Dong, J.; and Lam, K.-M. 2022.
\newblock Normattention-psn: A High-frequency Region Enhanced Photometric Stereo Network with Normalized Attention.
\newblock \emph{International Journal of Computer Vision}, 130(12): 3014--3034.

\bibitem[{Karaçalı and Snyder(2003)}]{tree}
Karaçalı, B.; and Snyder, W. 2003.
\newblock Reconstructing discontinuous surfaces from a given gradient field using partial integrability.
\newblock \emph{Computer Vision and Image Understanding}, 92(1): 78--111.

\bibitem[{Kim, Jung, and Lee(2024)}]{ae}
Kim, H.; Jung, Y.; and Lee, S. 2024.
\newblock Discontinuity-preserving Normal Integration with Auxiliary Edges.
\newblock In \emph{Proceedings of the IEEE/CVF Conference on Computer Vision and Pattern Recognition}, 11915--11923.

\bibitem[{Kossaifi et~al.(2023)Kossaifi, Kovachki, Azizzadenesheli, and Anandkumar}]{tfno}
Kossaifi, J.; Kovachki, N.; Azizzadenesheli, K.; and Anandkumar, A. 2023.
\newblock Multi-grid Tensorized Fourier Neural Operator for High-resolution PDEs.
\newblock \emph{arXiv:2310.00120}.

\bibitem[{Kovesi(2005)}]{shapelets}
Kovesi, P. 2005.
\newblock Shapelets Correlated with Surface Normals Produce Surfaces.
\newblock In \emph{Proceedings of the IEEE International Conference on Computer Vision}, volume~2, 994--1001.

\bibitem[{Li et~al.(2020{\natexlab{a}})Li, Kovachki, Azizzadenesheli, Liu, Bhattacharya, Stuart, and Anandkumar}]{fno}
Li, Z.; Kovachki, N.; Azizzadenesheli, K.; Liu, B.; Bhattacharya, K.; Stuart, A.; and Anandkumar, A. 2020{\natexlab{a}}.
\newblock Fourier Neural Operator for Parametric Partial Differential Equations.
\newblock arXiv:2010.08895.

\bibitem[{Li et~al.(2020{\natexlab{b}})Li, Kovachki, Azizzadenesheli, Liu, Bhattacharya, Stuart, and Anandkumar}]{no}
Li, Z.; Kovachki, N.; Azizzadenesheli, K.; Liu, B.; Bhattacharya, K.; Stuart, A.; and Anandkumar, A. 2020{\natexlab{b}}.
\newblock Neural Operator: Graph Kernel Network for Partial Differential Equations.
\newblock arXiv:2003.03485.

\bibitem[{Lichy, Sengupta, and Jacobs(2022)}]{lichy}
Lichy, D.; Sengupta, S.; and Jacobs, D.~W. 2022.
\newblock Fast Light-Weight Near-Field Photometric Stereo.
\newblock In \emph{Proceedings of the IEEE/CVF Conference on Computer Vision and Pattern Recognition}, 12602--12611.

\bibitem[{Lichy et~al.(2021)Lichy, Wu, Sengupta, and Jacobs}]{recnet}
Lichy, D.; Wu, J.; Sengupta, S.; and Jacobs, D.~W. 2021.
\newblock Shape and Material Capture at Home.
\newblock In \emph{Proceedings of the IEEE/CVF Conference on Computer Vision and Pattern Recognition}, 6119--6129.

\bibitem[{Liu, Jafarzadeh, and Yu(2023)}]{dafno}
Liu, N.; Jafarzadeh, S.; and Yu, Y. 2023.
\newblock Domain Agnostic Fourier Neural Operators.
\newblock In \emph{Proceedings of the Advances in Neural Information Processing Systems}, volume~36, 47438--47450. Curran Associates, Inc.

\bibitem[{Logothetis, Mecca, and Cipolla(2017)}]{logothetis2017semi}
Logothetis, F.; Mecca, R.; and Cipolla, R. 2017.
\newblock Semi-calibrated near field photometric stereo.
\newblock In \emph{Proceedings of the IEEE Conference on Computer Vision and Pattern Recognition}, 941--950.

\bibitem[{Loshchilov and Hutter(2017)}]{adamw}
Loshchilov, I.; and Hutter, F. 2017.
\newblock Decoupled Weight Decay Regularization.
\newblock \emph{arXiv:1711.05101}.

\bibitem[{Lu et~al.(2021)Lu, Jin, Pang, Zhang, and Karniadakis}]{deeponet}
Lu, L.; Jin, P.; Pang, G.; Zhang, Z.; and Karniadakis, G.~E. 2021.
\newblock {Learning Nonlinear Operators via DeepONet Based on the Universal Approximation Theorem of Operators}.
\newblock \emph{Nature Machine Intelligence}, 3(3): 218--229.

\bibitem[{Luo et~al.(2016)Luo, Li, Urtasun, and Zemel}]{erf}
Luo, W.; Li, Y.; Urtasun, R.; and Zemel, R. 2016.
\newblock Understanding the Effective Receptive Field in Deep Convolutional Neural Networks.
\newblock In \emph{Proceedings of the Advances in Neural Information Processing Systems}, 4905–4913.

\bibitem[{Mecca et~al.(2021)Mecca, Logothetis, Budvytis, and Cipolla}]{luces}
Mecca, R.; Logothetis, F.; Budvytis, I.; and Cipolla, R. 2021.
\newblock Luces: A dataset for Near-field Point Light Source Photometric Stereo.
\newblock arXiv:2104.13135.

\bibitem[{Ming et~al.(2021)Ming, Meng, Fan, and Yu}]{mono}
Ming, Y.; Meng, X.; Fan, C.; and Yu, H. 2021.
\newblock Deep Learning for Monocular Depth Estimation: A Review.
\newblock \emph{Neurocomputing}, 438: 14--33.

\bibitem[{Morton and Mayers(2005)}]{morton2005numerical}
Morton, K.~W.; and Mayers, D.~F. 2005.
\newblock \emph{Numerical Solution of Partial Differential Equations: An Introduction}.
\newblock Cambridge university press.

\bibitem[{Ouyang, Wang, and Chen(2022)}]{ir2}
Ouyang, H.; Wang, T.; and Chen, Q. 2022.
\newblock Restorable Image Operators with Quasi-Invertible Networks.
\newblock \emph{Proceedings of the AAAI Conference on Artificial Intelligence}, 36(2): 2008--2016.

\bibitem[{Paszke et~al.(2017)Paszke, Gross, Chintala, Chanan, Yang, DeVito, Lin, Desmaison, Antiga, and Lerer}]{pytorch}
Paszke, A.; Gross, S.; Chintala, S.; Chanan, G.; Yang, E.; DeVito, Z.; Lin, Z.; Desmaison, A.; Antiga, L.; and Lerer, A. 2017.
\newblock Automatic differentiation in pytorch.

\bibitem[{Patil et~al.(2022)Patil, Sakaridis, Liniger, and Van~Gool}]{plane2}
Patil, V.; Sakaridis, C.; Liniger, A.; and Van~Gool, L. 2022.
\newblock P3Depth: Monocular Depth Estimation with a Piecewise Planarity Prior.
\newblock In \emph{Proceedings of the IEEE/CVF Conference on Computer Vision and Pattern Recognition}, 1600--1611.

\bibitem[{Peng et~al.(2021)Peng, Liao, Wong, Luo, Zhou, and Chellappa}]{ct2}
Peng, C.; Liao, H.; Wong, G.; Luo, J.; Zhou, S.~K.; and Chellappa, R. 2021.
\newblock XraySyn: Realistic View Synthesis From a Single Radiograph Through CT Priors.
\newblock \emph{Proceedings of the AAAI Conference on Artificial Intelligence}, 35(1): 436--444.

\bibitem[{Qu{\'e}au and Durou(2015)}]{wls}
Qu{\'e}au, Y.; and Durou, J.-D. 2015.
\newblock Edge-Preserving Integration of a Normal Field: Weighted Least-Squares, TV and $L^1$ Approaches.
\newblock In \emph{Proceedings of Scale Space and Variational Methods in Computer Vision}, 576--588. Springer.

\bibitem[{Quéau, Durou, and Aujol(2017{\natexlab{a}})}]{survey}
Quéau, Y.; Durou, J.-D.; and Aujol, J.-F. 2017{\natexlab{a}}.
\newblock {Normal Integration: A survey}.
\newblock \emph{Journal of Mathematical Imaging and Vision}, 60(4): 576--593.

\bibitem[{Quéau, Durou, and Aujol(2017{\natexlab{b}})}]{variational}
Quéau, Y.; Durou, J.-D.; and Aujol, J.-F. 2017{\natexlab{b}}.
\newblock {Variational Methods for Normal Integration}.
\newblock \emph{Journal of Mathematical Imaging and Vision}, 60(4): 609--632.

\bibitem[{Raonic et~al.(2023)Raonic, Molinaro, De~Ryck, Rohner, Bartolucci, Alaifari, Mishra, and de~B\'{e}zenac}]{cno}
Raonic, B.; Molinaro, R.; De~Ryck, T.; Rohner, T.; Bartolucci, F.; Alaifari, R.; Mishra, S.; and de~B\'{e}zenac, E. 2023.
\newblock Convolutional Neural Operators for Robust and Accurate Learning of PDEs.
\newblock In \emph{Proceedings of the Advances in Neural Information Processing Systems}, volume~36, 77187--77200. Curran Associates, Inc.

\bibitem[{Shi et~al.(2019{\natexlab{a}})Shi, Mo, Wu, Duan, Yeung, and Tan}]{shi2019benchmark}
Shi, B.; Mo, Z.; Wu, Z.; Duan, D.; Yeung, S.-K.; and Tan, P. 2019{\natexlab{a}}.
\newblock A Benchmark Dataset and Evaluation for Non-Lambertian and Uncalibrated Photometric Stereo.
\newblock \emph{IEEE Transactions on Pattern Analysis and Machine Intelligence}, 41(02): 271--284.

\bibitem[{Shi et~al.(2019{\natexlab{b}})Shi, Mo, Wu, Duan, Yeung, and Tan}]{diligent}
Shi, B.; Mo, Z.; Wu, Z.; Duan, D.; Yeung, S.-K.; and Tan, P. 2019{\natexlab{b}}.
\newblock A Benchmark Dataset and Evaluation for Non-Lambertian and Uncalibrated Photometric Stereo.
\newblock \emph{IEEE Transactions on Pattern Analysis and Machine Intelligence}, 41(2): 271--284.

\bibitem[{Simchony, Chellappa, and Shao(1990)}]{dct}
Simchony, T.; Chellappa, R.; and Shao, M. 1990.
\newblock Direct Analytical Methods for Solving Poisson Equations in Computer Vision Problems.
\newblock \emph{IEEE Transactions on Pattern Analysis and Machine Intelligence}, 12(5): 435--446.

\bibitem[{Sofiiuk et~al.(2020)Sofiiuk, Petrov, Barinova, and Konushin}]{mask}
Sofiiuk, K.; Petrov, I.; Barinova, O.; and Konushin, A. 2020.
\newblock F-BRS: Rethinking Backpropagating Refinement for Interactive Segmentation.
\newblock In \emph{Proceedings of the IEEE/CVF Conference on Computer Vision and Pattern Recognition}, 8620--8629.

\bibitem[{Wang, Xie, and Cui(2020)}]{DGP3}
Wang, M.; Xie, W.; and Cui, M. 2020.
\newblock Surface Reconstruction with Unconnected Normal Maps: An Efficient Mesh-based Approach.
\newblock In \emph{Proceedings of the ACM International Conference on Multimedia}, 2617–2625. Association for Computing Machinery.

\bibitem[{Woodham(1980)}]{ps}
Woodham, R.~J. 1980.
\newblock {Photometric Method for Determining Surface Orientation from Multiple Images}.
\newblock \emph{Optical Engineering}, 19(1): 191139.

\bibitem[{Wu and Li(1988)}]{early}
Wu, Z.; and Li, L. 1988.
\newblock A Line Integration Based Method for Depth Recovery from Surface Normals.
\newblock In \emph{Proceedings of International Conference on Pattern Recognition}, 591--595.

\bibitem[{Xie et~al.(2019)Xie, Wang, Wei, Jiang, and Qin}]{DGP2}
Xie, W.; Wang, M.; Wei, M.; Jiang, J.; and Qin, J. 2019.
\newblock Surface Reconstruction From Normals: A Robust DGP-Based Discontinuity Preservation Approach.
\newblock In \emph{Proceedings of the IEEE/CVF Conference on Computer Vision and Pattern Recognition}, 5323--5331.

\bibitem[{Xie et~al.(2014)Xie, Zhang, Wang, and Chung}]{DGP}
Xie, W.; Zhang, Y.; Wang, C.~C.; and Chung, R. C.-K. 2014.
\newblock Surface-from-Gradients: An Approach Based on Discrete Geometry Processing.
\newblock In \emph{Proceedings of the IEEE/CVF Conference on Computer Vision and Pattern Recognition}, 2203--2210.

\bibitem[{Xu et~al.(2023)Xu, Wang, Ding, and Yang}]{igev}
Xu, G.; Wang, X.; Ding, X.; and Yang, X. 2023.
\newblock Iterative Geometry Encoding Volume for Stereo Matching.
\newblock In \emph{Proceedings of the IEEE/CVF Conference on Computer Vision and Pattern Recognition}, 21919--21928.

\bibitem[{Yang et~al.(2022)Yang, Yao, Huang, Zhou, and Zhao}]{yang2022sir}
Yang, Z.; Yao, M.; Huang, J.; Zhou, M.; and Zhao, F. 2022.
\newblock SIR-Former: Stereo Image Restoration Using Transformer.
\newblock In \emph{Proceedings of the 30th ACM International Conference on Multimedia}, 6377--6385.

\bibitem[{Zhang, Zhang, and Zhang(2020)}]{plane1}
Zhang, W.; Zhang, W.; and Zhang, Y. 2020.
\newblock GeoLayout: Geometry Driven Room Layout Estimation Based on Depth Maps of Planes.
\newblock In \emph{Proceedings of the European Conference on Computer Vision}, 632--648.

\bibitem[{Zhang(2000)}]{calibration}
Zhang, Z. 2000.
\newblock A Flexible New Technique for Camera Calibration.
\newblock \emph{IEEE Transactions on Pattern Analysis and Machine Intelligence}, 22(11): 1330--1334.

\bibitem[{Zhu and Smith(2020)}]{zs}
Zhu, D.; and Smith, W. A.~P. 2020.
\newblock Least Squares Surface Reconstruction on Arbitrary Domains.
\newblock In \emph{Proceedings of the European Conference on Computer Vision}, 530--545. Springer.

\bibitem[{Zisserman and Wiles(2017)}]{statue}
Zisserman, A.; and Wiles, O. 2017.
\newblock SilNet: Single-and Multi-View Reconstruction by Learning from Silhouettes.
\newblock In \emph{Proceedings of the British Machine Vision Conference}. British Machine Vision Association and Society for Pattern Recognition.

\end{thebibliography}

\clearpage

\setcounter{secnumdepth}{1} %May be changed to 1 or 2 if section numbers are desired.

\section{Supplementary Material}

In this supplementary material, we include more details presented in the main paper. In section \ref{pre}, we review the detailed mathematics problem description and our coordinate system to avoid scale ambiguity. In section \ref{relations}, we discuss more relations of our work to the previous especially for discontinuity preservation. In section \ref{discontinuity}, details of discontinuity and optimization is presented including expanded formula for the relative weight to model discontinuity. In section \ref{arch} and \ref{implementation}, we present the details and pseudo code for our network architecture as well as other baseline methods. In section \ref{additional}, we launch more experiments to show the performance of our approach under real-world Photometric Stereo settings. In section \ref{more_ablation}, we present more ablation details and experiments to emphasize properties for our neural operator. At last, in section \ref{limitation}, three limitations are clarified for future improvements.

\section{Preliminaries}
\label{pre}

\subsection{Problem Description}

In our SfG problem, consider $\textbf{x}=[x(u,v),y(u,v),z(u,v)]^T\in \mathbb{R}^3$, a point in the 3D space, is projected to the image plane with coordinates $\textbf{u}=[u,v]^T \in \mathbb{R}^2$ with its outer normal being $\textbf{n}(u,v)=[n_1(u,v),n_2(u,v),n_3(u,v)]^T\in \mathbb{R}^3$. The ultimate goal is to solve the following unified partial differentiation equations 

\begin{equation}
	\nabla \tilde{z}=[p,q]^T,
	\label{equation1}
\end{equation} 
where $\tilde{z}=z$, 
\begin{equation*}
	\begin{matrix}  
		p=-\frac{n_1}{n_3}, \\ 
		q=-\frac{n_2}{n_3},
	\end{matrix}
\end{equation*}
for orthographic projection and $\tilde{z}=ln(z)$, 
\begin{equation}
	\begin{matrix}  
		p=-\frac{n_1}{\tilde{n_3}}=-\frac{n_1}{un_1+vn_2+fn_3}, \\ 
		q=-\frac{n_2}{\tilde{n_3}}=-\frac{n_2}{un_1+vn_2+fn_3},
	\end{matrix}
	\label{perspective}
\end{equation}
for perspective projection. Ideally, when the surface is smooth and normal is accurate, the solution should be independent from the
integration path. In practice, discontinuity, and noise will sever disturbance. Besides, due to the Neumann boundary condition, the solution is non-unique \cite{survey}.

\subsection{Coordinate systems}

Standard pinhole camera model centered at the origin in world coordinates with z-axis pointing at the viewing direction can be formulated as 

\begin{equation}
	(u,v,1)^T\sim K(x,y,z)^T,
\end{equation} 
where $K$ is the camera  intrinics matrix. Then, a 3D point can be obtained by a depth map $z(u,v)$ through

\begin{equation}
	\textbf{x}(u,v) = z(u,v)K^{-1}(u,v,1)^T,
	\label{camera}
\end{equation}
and its normal map can be computed by 

\begin{equation}
	\textbf{n}(\textbf{x})=\frac{\frac{\partial \textbf{x}}{\partial u}\times\frac{\partial \textbf{x}}{\partial v}}{||\frac{\partial \textbf{x}}{\partial u}\times\frac{\partial \textbf{x}}{\partial v}||}.
	\label{nfd}
\end{equation}

However, if we solve Eq. \ref{equation1} with Eq. \ref{perspective} to obtain absolute depth directly, scale ambiguity will occur due to different main distance. Thus, we follow \cite{lichy} to compute relative depth where depth map $z(u,v)$ is replaced by $z(u,v)/\mu$ where $\mu$
is the main distance. This transformation also occur in other scale-ambiguous tasks \cite{mono}. We also vary intrinics $K$ to  constrain the image coordinate in $[-1,1]$ to avoid varying image size and perform further positional encoding in neural operator. In theory, our coordinate systems can handle any camera intrinics and resolution. Meanwhile, the point cloud of depth map is calculated using Eq.\ref{camera} and the normal $\textbf{n}_i(\textbf{x})$ as well as $\textbf{n}_i(\textbf{x}^{gt})$ from loss function in the main paper 

\begin{equation}
	\mathcal{L}_r=\frac{1}{N}\sum_{i=1}^N((1-\omega_i)||z_i-z^{gt}_i||_1+\gamma \omega_i||\textbf{n}_i(\textbf{x}) - \textbf{n}_i(\textbf{x}^{gt})||_1).
	\label{attention_weighted}
\end{equation}
is computed using Eq.\ref{nfd}.

\section{More Relations to previous work}
\label{relations}
In this section, we will discuss more on the relations between previous optimization based methods and our FNIN.

The idea of using optimization is the milestone in SfG problem. Since Horn and Brooks \shortcite{hb} suggested to minimize the difference between
input $\textbf{g}$ and the gradient of depth $z$ by

\begin{equation}
	F = \iint \limits_{(u,v)\in D} ||\nabla z(u,v) - \textbf{g}(u,v)||^2dudv,
	\label{quadratic}
\end{equation}

Many approaches have been proposed to strengthen robustness, including enforcing integrability \cite{fft, dct, wavelet, shapelets}, discrete geometry processing \cite{DGP, DGP2, DGP3, ipf}, and robust estimator \cite{triple, variational, zs, wls}. Although being a learning method, our FNIN draws inspiration from many previous methods to utilize their advantages.

For discontinuity optimization, an impactful work is the introduction of weighted least square (WLS) functional by  Qu\'eau and Durou \shortcite{wls},

\begin{equation}
	\begin{aligned}
		F = \iint \limits_{(u,v)\in D} &\omega(u,v)||\nabla z(u,v) - \textbf{g}(u,v)||^2 \\
		+&\frac{\lambda}{2}(z(u,v)-z_0(u,v))^2dudv.
	\end{aligned}
	\label{wls}
\end{equation}
where $\omega(u,v)$ is prior weight and $z_0(u,v)$ is prior depth. Initially, the weight and prior depth are estimated using additional information, but they fail to combine depth information to weight accurately , which is noticed by recent works \cite{bini,ae} and ours.

\section{Discontinuity Detection \& Optimization}
\label{discontinuity}
In this section, we will describe several details in our attention net for discontinuity detection.

Our objective 
\begin{equation}
	\begin{aligned}
		\mathop{\min}_{z(u,v)} \sum_{D} &\frac{w_r}{2}(n_3\partial_u^+z+n_1)^2+\frac{w_l}{2}(n_3\partial_u^-z+n_1)^2\\
		+&\frac{w_t}{2}(n_3\partial_v^+z+n_2)^2+\frac{w_b}{2}(n_3\partial_v^-z+n_2)^2\\
		+&\lambda(z(u,v)-z_R(u,v))^2,
	\end{aligned}
	\label{our_objective}
\end{equation}
in the main paper is the natural expression of Eq. \ref{wls}. To generate the relative weight through our attention net

\begin{equation}
	\begin{matrix}
		\Phi = f_{\theta_{ae}}(\Delta z;\theta_{ae}),\\
		\omega = \mathrm{normalize}( f_{\theta_{ar}}(\Phi;\theta_{ar})),
	\end{matrix}
\end{equation}
we model discontinuity using $dis=(\Delta_u^+z)^2/\delta_u+(\Delta_u^-z)^2/\delta_u+(\Delta_v^+z)^2/\delta_v+(\Delta_v^-z)^2/\delta_v$, $\delta_u$ and $\delta_v$ are the discretization size in horizontal and vertical direction respectively to adapt to different resolution. $\Delta_u^- z, \Delta_u^+ z, \Delta_v^- z, \Delta_v^+ z$ are the  difference between one-side limit and its value from four directions of $\Delta z$, which is defined as 
\begin{equation}
	\begin{matrix}
		\Delta_u^+z\approx n_z(z(u,v)-z(u+1, v)),\\ 
  \Delta_u^-z\approx n_z(z(u,v)-z(u-1, v)),\\
		\Delta_v^+z\approx n_z(z(u,v)-z(u, v+1)),\\
  \Delta_v^-z\approx n_z(z(u,v)-z(u, v-1)).\\
	\end{matrix}
	\label{one-side}
\end{equation}

\section{Network Architecture}
\label{arch}
Generally, the proposed Fourier neural operator based numerical integration network has three major components: initial net, iterative net and attention net. In our implementation, initial net and iterative net share the same structure, both of them are composed by 4 Fourier layers. We use GELU function for activation and lifting operation $P$ as well as projection operation $Q$ are performed by pixel-wise MLP. Positional encoding uses our coordinate system. Illustrated in the main paper, one forward pass of FNIN is presented in Algorithm \ref{alg1}.

\begin{algorithm}[h]
	\caption{Forward pass of FNIN}\label{forward}
	\begin{algorithmic}
		\STATE \textbf{input:} surface normal $\textbf{n}(u,v)$, camera intrinsics $K$.
		\STATE \textbf{output:} depth map $z_R$, discontinuity map $\omega_R$.
		\STATE 1: $\textbf{n}_1$,...,$\textbf{n}_R$ = \textit{Downsample}($\textbf{n}(u,v)$);
		\STATE 2: initialize $z_0$;
		\STATE 3: \textbf{for} r = 1 \textbf{to} R \textbf{do}
		\STATE 4: \hspace{0.5cm} $\hat{z}=ln(z_{i-1})$; //log space
		\STATE 5: \hspace{0.5cm} $\nabla\hat{z}=(\hat{z}_{(m+1)n}-\hat{z}_{(m-1)n}, \hat{z}_{m(n+1)}-\hat{z}_{m(n-1)})$; //central difference
		\STATE 6: \hspace{0.5cm} compute $\textbf{g}=(p_i, q_i)$ using perspective correction in Eq.\ref{perspective};
		\STATE 7: \hspace{0.5cm} $v_0=P(\textbf{g}-\nabla\hat{z})$; //lifting to high dimensional representation.
		\STATE 8: \hspace{0.5cm} \textbf{for} t = 0 \textbf{to} T-1 \textbf{do}
		\STATE 9: \hspace{1cm} \textbf{if} r = 1 \textbf{then}
		\STATE 10: \hspace{1.5cm} use $W_t $ and $\theta_t$ from initial net;
		\STATE 11: \hspace{1cm} \textbf{else}
		\STATE 12: \hspace{1.5cm} use $W_t $ and $\theta_t$ from iterative net;
		\STATE 13: \hspace{1cm} \textbf{end if}
		\STATE 14: \hspace{1cm} $v_{t+1}=\sigma(W_tv_t+F^{-1}(R_{\theta_t}(Fv_t)))$; //update 
		\STATE 15: \hspace{0.5cm} \textbf{end for}
		\STATE 16: \hspace{0.5cm} $dis=(\Delta_u^+z)^2/\delta_u+(\Delta_u^-z)^2/\delta_u+(\Delta_v^+z)^2/\delta_v+(\Delta_v^-z)^2/\delta_v$
		\STATE 17: \hspace{0.5cm} $\Phi = f_{\theta_{ae}}(dis;\theta_{ae})$ 
		\STATE 18: \hspace{0.5cm} $\omega=\textit{normalize}( f_{\theta_{ar}}(\Phi;\theta_{ar}))$
		\STATE 19: \hspace{0.5cm} $z_{r+1}= $ \textit{Upsample}(\textit{exp}(Q($v_T$)))
		\STATE 20: \textbf{end for}
		
	\end{algorithmic}
	\label{alg1}
\end{algorithm}


\section{Implementation Details}
\label{implementation}
In this section, we will present the training details of our approach as well as the parameter settings in baseline methods.

 During the training process, we use AdamW \cite{adamw} optimizer and exponential learning rate schedule with $\gamma=0.9$. Final models are trained 15 epochs with a maximum learning rate of $2e^{-3}$ with a batch size of 20. As for data augmentation, we randomly zero $50\times 50$ patches and add random noise following \cite{lichy}. The entire training process finishes within 5 hours.

For baseline methods, we set $k=3,m=25,\lambda=1$ in ZS \cite{zs}, $\alpha=1e^{-1}, \lambda=1e^{-4}$ in WLS \cite{wls}, $\alpha=1e^{-3}, \lambda=1e^{-5}$ in TV \cite{variational}, $\mu=0.2,\nu=10$ in AD \cite{variational}, $\mu=45$ in MS \cite{variational}, $k=2$ in BiNI \cite{bini}. For Rec-Net \cite{lichy}, we follow their training process and set $\lambda=1$.

\begin{figure*}[t]	
\centering
\includegraphics[width=\linewidth]{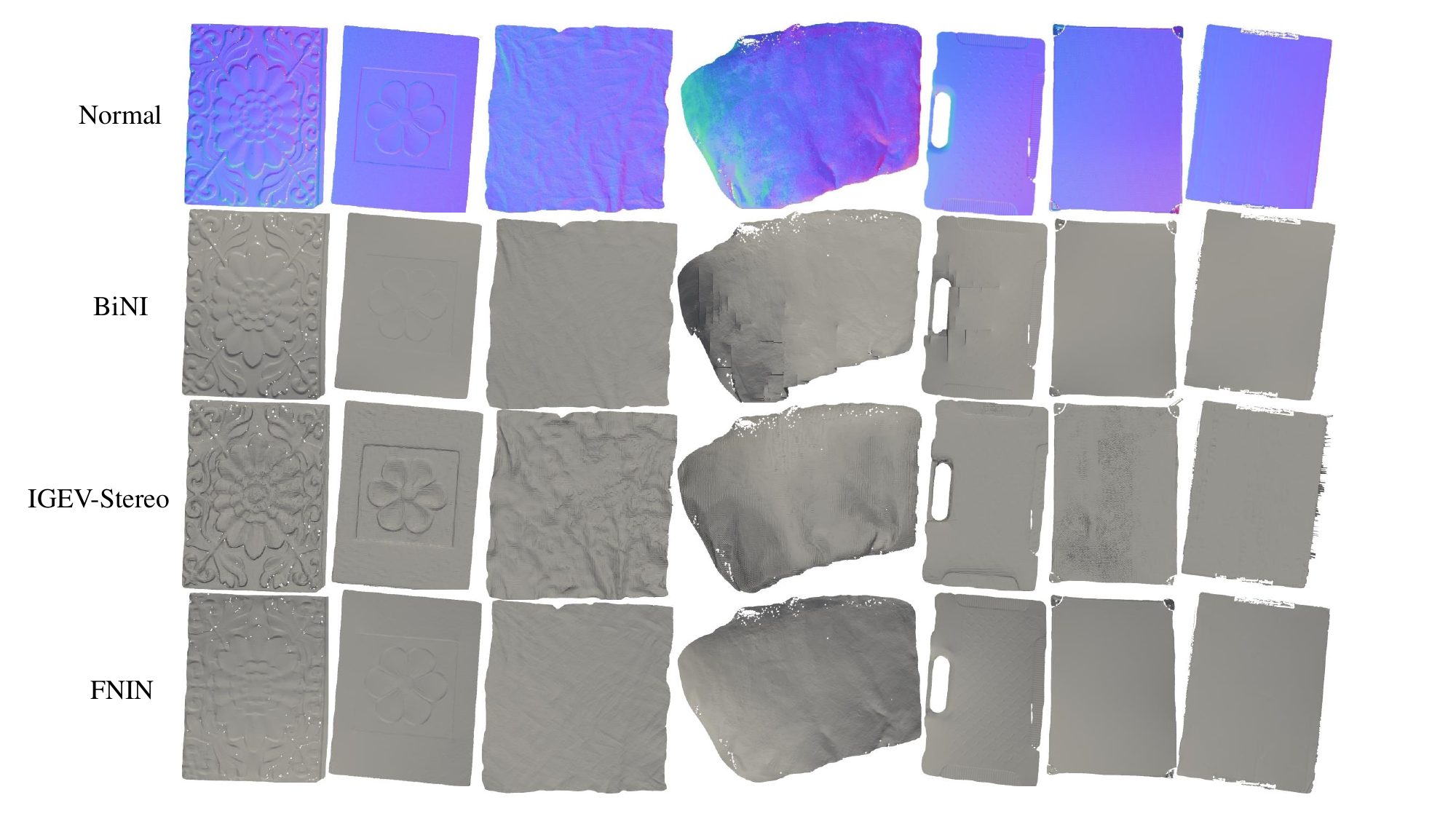}
	\caption{Visual results on our dataset. Input normal estimated by classic PS \cite{ps}, mesh estimated by BiNI \cite{bini}, IGEV-Stereo \cite{igev} and our FNIN are illustrated respectively.}
	\label{ps}  
\end{figure*}

\section{Additional Experiment}
\label{additional}
\subsection{Image capture}

\begin{figure}[h]    
	\centering
	\begin{minipage}[c]{0.45\linewidth}
		\subfigure[Image capture device]{
			\includegraphics[height=1.82in]{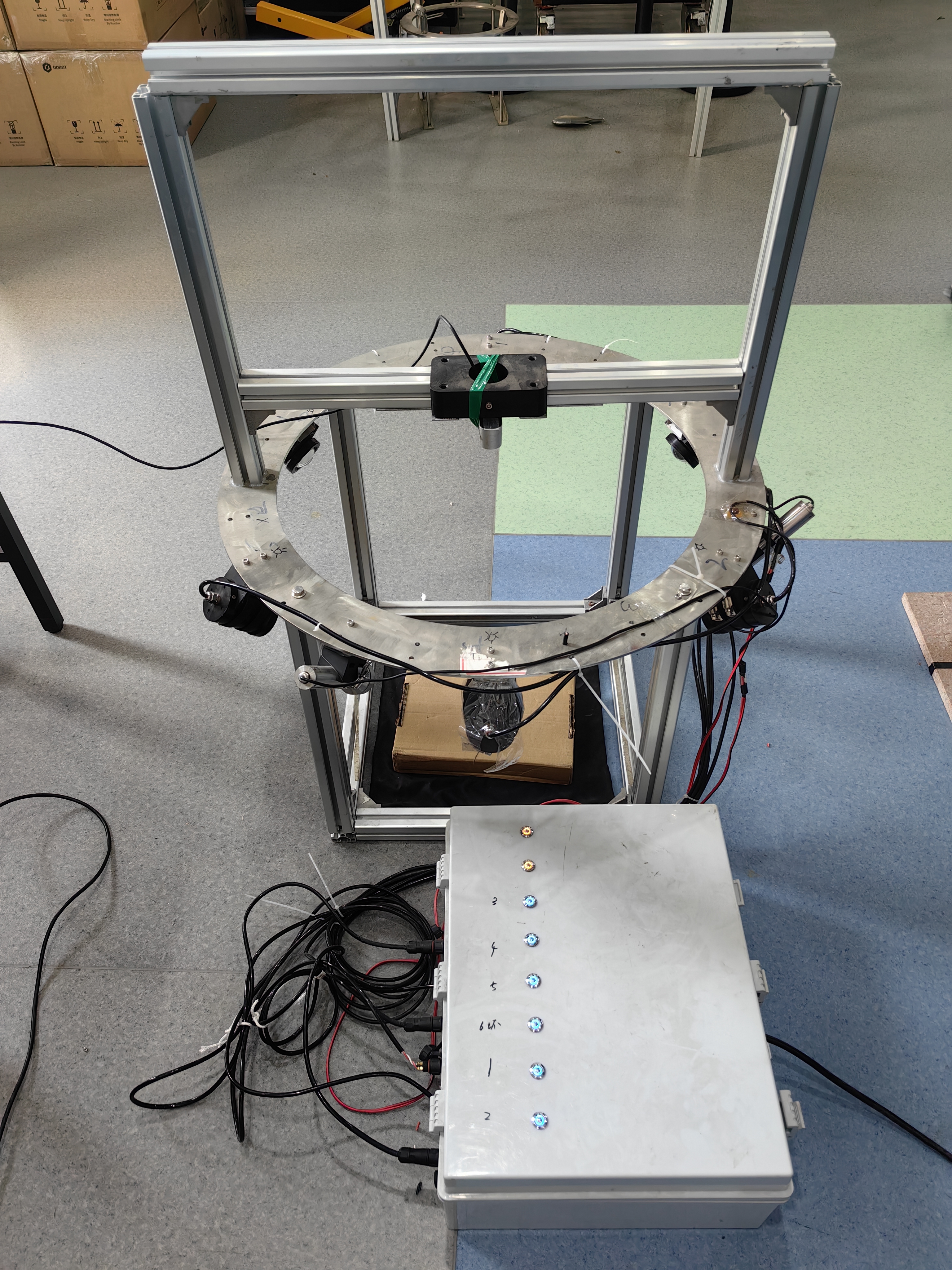}} 
	\end{minipage}
	\begin{minipage}[c]{0.45\linewidth}
		\centering
		\subfigure[left]{
			\includegraphics[height=0.75in]{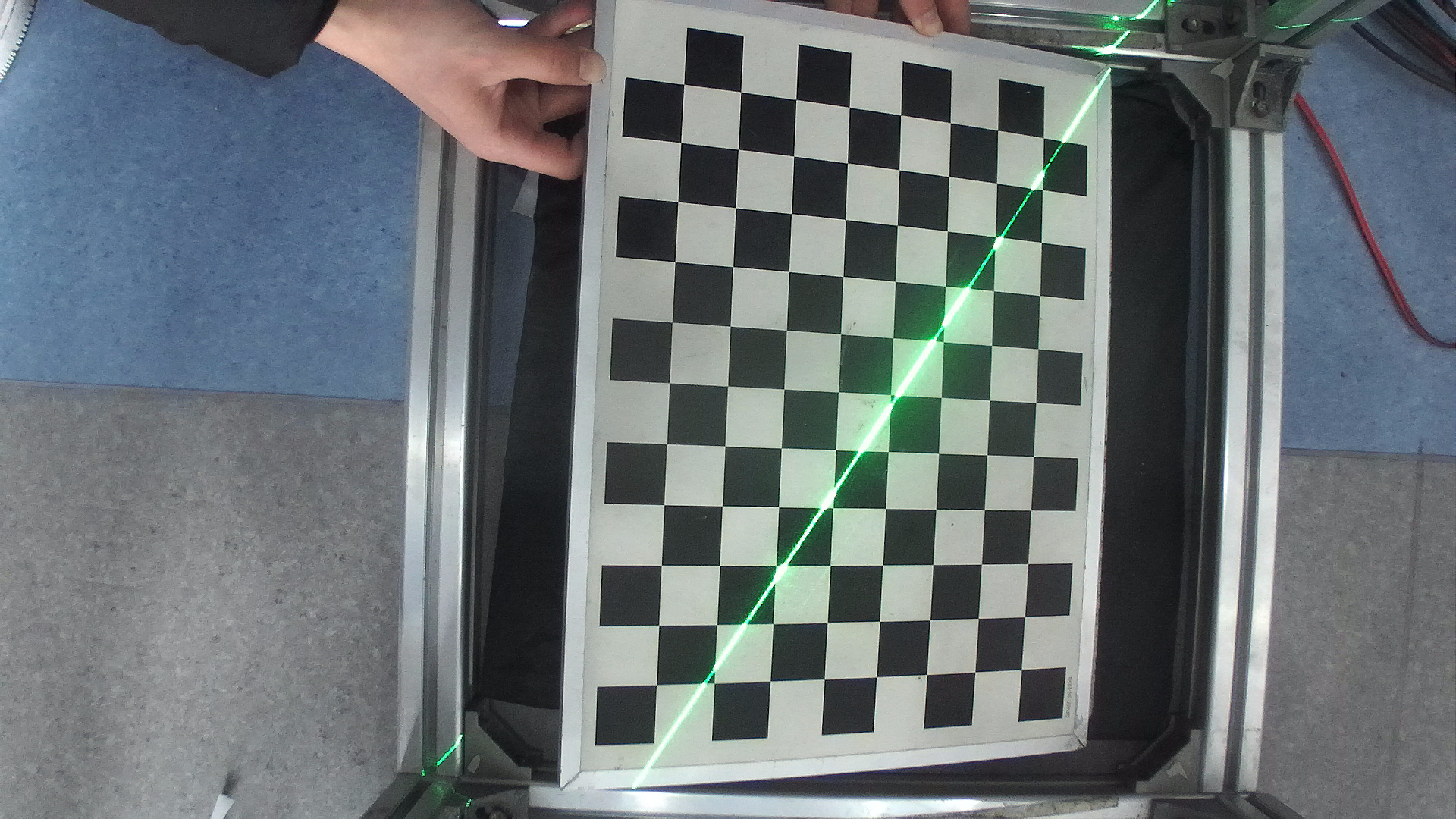}}	\\
		\subfigure[right]{
			\includegraphics[height=0.75in]{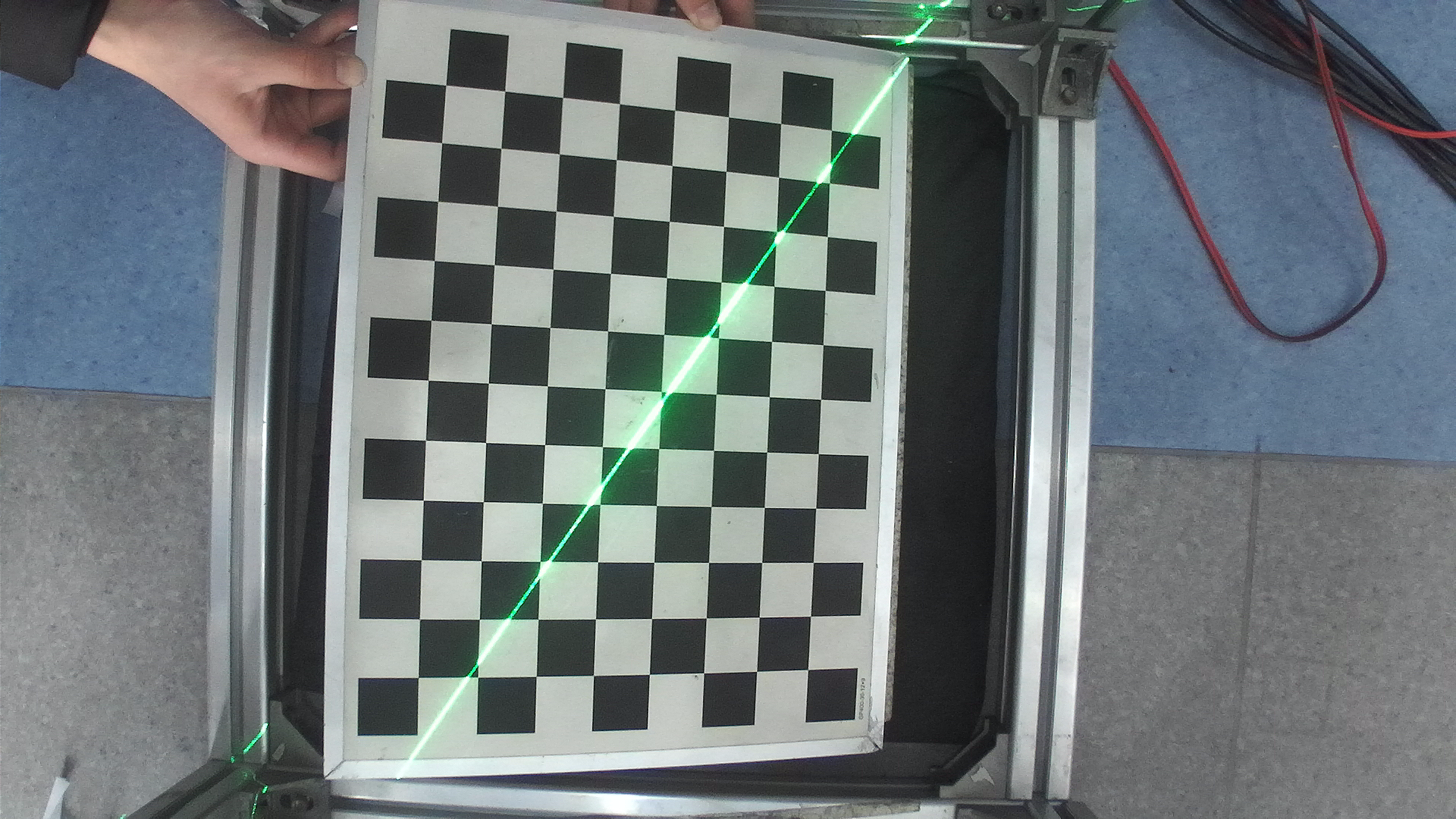}}\\
	\end{minipage}
	\caption{Image capture device and one pair of calibration images.} 
	\label{equip}            
\end{figure}

To demonstrate the performance of our method under real-world Photometric Stereo settings, we captured a new dataset with abundant textures containing 7 objects. We provide binocular images for each of the object with cameras placed in the center of the device and 6 light sources with identical brightness are irradiated from $0^{\circ}, 60^{\circ}, 120^{\circ}, 180^{\circ}, 240^{\circ}, 300^{\circ}$ respectively. Camera intrinsics are obtained based on stereo camera calibration \cite{calibration}. Additionally, we used interactive segmentation tools \cite{mask} to generate masks. Image capture device and one pair of calibration images are shown in Fig. \ref{equip}.

\subsection{Visual results}

Input normal maps are computed using classic Photometric Stereo \cite{ps} where the reflection intensity $\mathbf{I}$ is linearly proportional to the angle between the normal $\mathbf{n}$ and incident light $\mathbf{l}$ with

\begin{equation}
    \mathbf{I} \propto \mathbf{l}^T \mathbf{n}
\end{equation}

We illustrate the results computed by BiNI \cite{bini} and our FNIN. Additionally, we estimate the depth through advanced stereo matching network IGEV-Stereo \cite{igev} using rectified binocular images. Through qualitative comparison, our FNIN gives the most reasonable estimation while BiNI develops unexpected jumps and IGEV-Stereo cannot preserve details accurately.

\begin{table}[t]
	\centering
	\setlength{\tabcolsep}{0.75mm}{
		\begin{tabular}{ccc}
			\toprule
			n\_layer &MAE on DiLiGenT&MAE on LUCES \\
			\midrule
			1&6.38&4.40\\
                10&6.03&2.93\\	
                4 (Final)&5.91&3.00\\

			\bottomrule
		\end{tabular}
	}
    \caption{Ablation on the number of layers. \label{layers}}
\end{table}

\begin{table}[t]
	\centering
	\setlength{\tabcolsep}{0.75mm}{
		\begin{tabular}{ccc}
			\toprule
			n\_layer &MAE on DiLiGenT&MAE on LUCES \\
			\midrule
			4&8.11&4.83\\
                8&6.32&3.51\\	
                16 (Final)&5.91&3.00\\

			\bottomrule
		\end{tabular}
	}
    \caption{Ablation on the truncated mode. \label{mode}}
\end{table}

\section{Ablation}
\label{more_ablation}

\subsection{More Details}
As traditional optimization based methods doesn't depend on randomness, in this section, we will only mainly focus on the statistical comparison in ablation study. All experiments are repeated on five trials with five different random seeds.

\begin{table*}[!t]
	\caption{Mean and standard deviation of the errors. Discontinuity optimization is not performed in experiments with *. \ding{56} indicates a numerical failure.  \label{statistic}}
	\centering

	\setlength{\tabcolsep}{1.5mm}{
		\begin{tabular}{cccccc}
			\toprule%第一道横线
			  Experiments & w/o attention$^*$ & w/ mask concatenation$^*$ & w/o discontinuity optimization &Rec-Net & FNIN\\
			
			\midrule%第二道横线 
			DiLiGenT (mm) & $5.97\pm 0.35$& $17.78\pm 2.26$& $5.83\pm 0.19$ & \ding{56} & $4.5456\pm 0.0083$\\
   
			LUCES (mm) & $4.20\pm 0.42$& $4.18\pm 0.42$& $3.05\pm 0.06$ & $6.13\pm 2.41$ & $1.9246\pm 0.0004$\\
			\bottomrule%第三道横线
		\end{tabular}
	}
\end{table*}

In Tab. \ref{statistic}, we provide mean and standard deviation of the errors. It is noteworthy that attention and mask concatenation will  increase not only the errors but also its standard deviation, because they will develop uncertain jumps at discontinuities. Additionally, Wilcoxon signed-rank tests between each ablation experiments and FNIN are obvious.

\subsection{Ablation on the number of Fourier layers}

In our implementation, we follow the original settings of FNO. Here we present experiments with different Fourier layers (Discontinuity optimization is removed for comparison) in  Tab. \ref{layers}. It's a trade off between accuracy and computation cost. The final setting is 4.

\subsection{Ablation on truncated mode}

The truncated mode determines the ability of approximating complex solution operator. In Tab. \ref{mode}, it's obvious that fewer modes will lead to a significant decrease in accuracy.

\section{Limitations}
\label{limitation}
\begin{figure}[H]	
\centering
\includegraphics[width=0.5\linewidth]{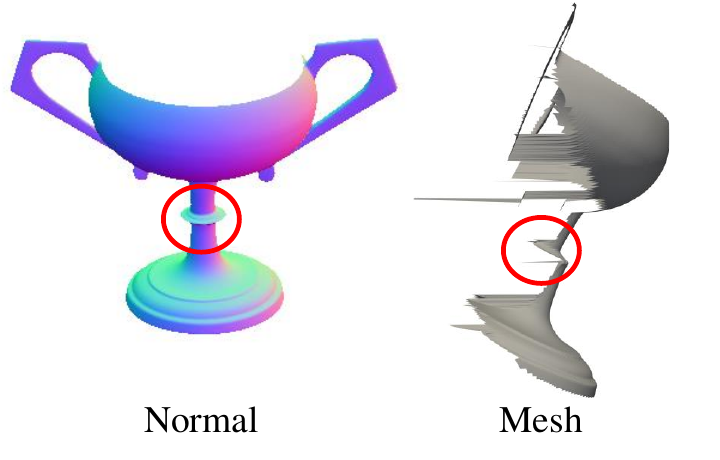}
	\caption{Surface islands.}
	\label{limitation1}  
\end{figure}

First, surface islands caused by scale or offset ambiguity still exists. Shown in Fig. \ref{limitation1}, discontinuities divide the domain into separate domains where the jumps at the edges cannot be implied by normal map. Traditional optimization based methods \cite{bini,ae} also encounter this issue, but in practice, our FNIN can easily combine with other deep learning algorithm to introduce correct scale or offset (e.g., piecewise planarity prior in monocular depth estimation \cite{plane1,plane2}).

\begin{figure}[H]	
\centering
\includegraphics[width=0.5\linewidth]{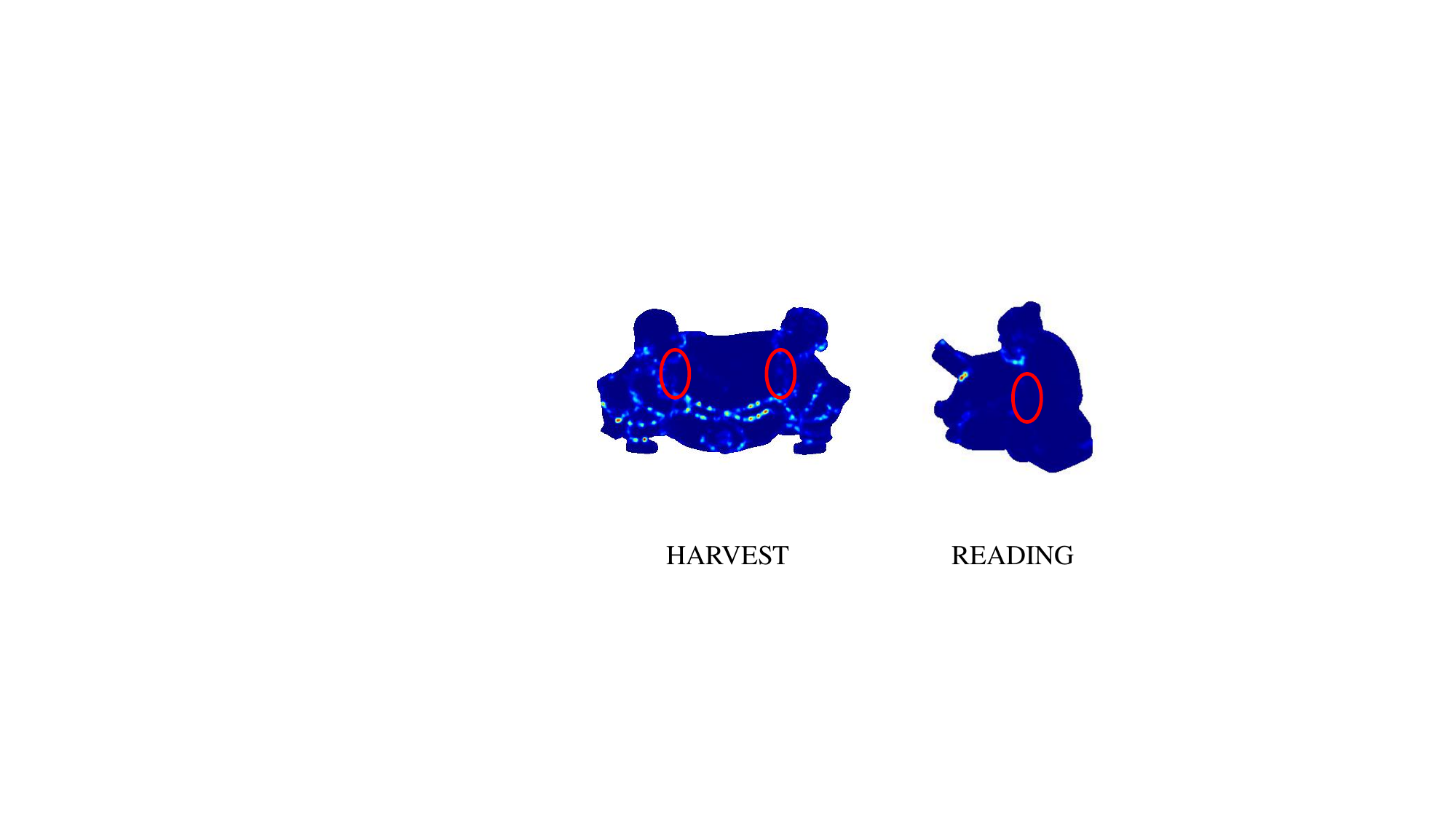}
	\caption{Insufficient weight.}
	\label{limitation2}  
\end{figure}

Second, shown in Fig. \ref{limitation2}, as we decouple discontinuity detection and its optimization, the intensity of relative weight for discontinuity relies on training data. However, due to the lack of details in synthetic data, attention to large discontinuity may not be enough. Therefore, proper regularization or customized training data would be desirable.

\begin{figure}[H]	
\centering
\includegraphics[width=0.5\linewidth]{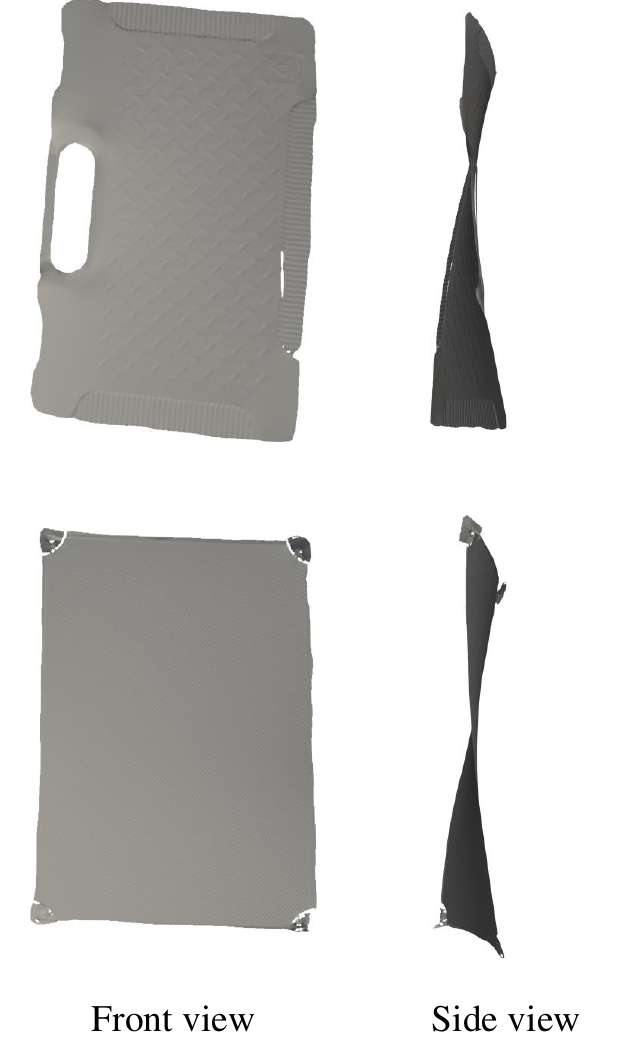}
	\caption{Near-field light bias.}
	\label{limitation3}  
\end{figure}

At last, extending our FNIN to nonlinear equation is still an open problem. Previous optimization based methods attempt to solve such equations in \textit{near-field} Photometric Stereo without normal estimation \cite{logothetis2017semi}, where further consideration to remove the dependence  between depth and gradient field is necessary. Shown in Fig. \ref{limitation3}, there is an obvious bias in the side view.

\end{document}